\RequirePackage{fix-cm}
\documentclass[twocolumn]{svjour3}          
\smartqed  

\usepackage{graphicx}

\usepackage{color}
\usepackage{mathptmx}      
\usepackage{array}
\usepackage{textcomp}
\usepackage[authoryear]{natbib}
\usepackage{caption}
\usepackage{enumitem}
\usepackage{url}

\usepackage{subfig}
\usepackage{enumitem}
\usepackage{ulem}
\newcommand{\TODO}[1]{{\textcolor[RGB]{23,165,83}{\textbf{[TODO: #1]}}}}
\newcommand{\REV}[1]{#1}

\newcommand{\REVV}[1]{#1}
\newcommand{\OLD}[1]{}
\newenvironment{REVregion}{}

\begin{document}
\captionsetup[figure]{labelfont={bf},labelformat={default},labelsep=space,name={Fig.}}
\title{Virtual Training for a Real Application: \\ Accurate Object-Robot Relative Localization without Calibration
}


\author{Vianney Loing   \and
        Renaud Marlet	\and
	Mathieu Aubry
}


\institute{
V. Loing is with
Navier (UMR 8205), Ecole des Ponts ParisTech, UPE, Champs-sur-Marne, France\\
             \email{vianney.loing@enpc.fr}     \\      
M. Aubry and R. Marlet are with
        LIGM (UMR 8049), Ecole des Ponts ParisTech, UPE, Champs-sur-Marne, France
}

\date{Received: date / Accepted: date}

\maketitle

\begin{abstract}
Localizing an object accurately with respect to a robot is a key step for autonomous robotic manipulation. In this work, we propose to tackle this task knowing only 3D models of the robot and object in the particular case where the scene is viewed from uncalibrated cameras --- a situation which would be typical in an uncontrolled environment, e.g., on a construction site. We demonstrate that this localization can be performed very accurately, with millimetric errors, without using a single real image for training, a strong advantage since acquiring representative training data is a long and expensive process. Our approach relies on a classification Convolutional Neural Network (CNN) trained using hundreds of thousands of synthetically rendered scenes with randomized parameters. To evaluate our approach quantitatively and make it comparable to alternative approaches, we build a new rich dataset of real robot images with accurately localized blocks.

\end{abstract}

\section{Introduction}
\label{intro}

\begin{figure}[htp]
  \centering
  \subfloat[Relative localization as classification, here with 5\,mm square bins (red grid) for coarse estimation, with overlaid 5\,cm squares (blue grid) to give a sense of dimension]{\label{fig:zones}\includegraphics[width=0.85\columnwidth]{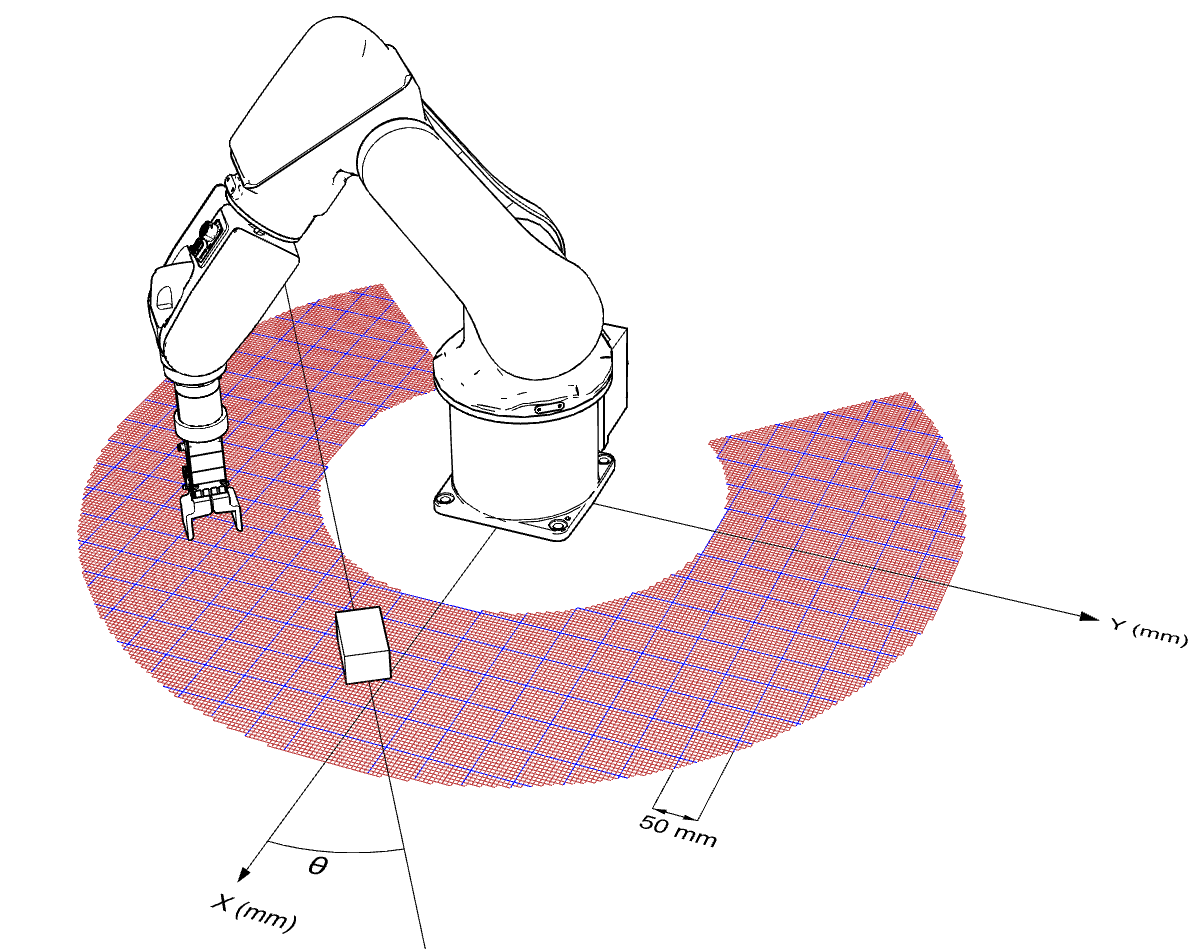}}
  \hspace{0pt}
  \subfloat[Training image]{\label{fig:trainimage}\includegraphics[height=0.45\columnwidth]{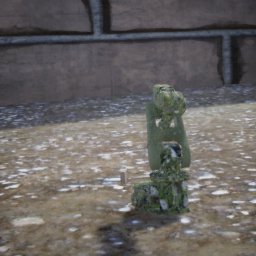} }
  \hspace{5pt}
  \subfloat[Test image]{\label{fig:testimage}\includegraphics[height=0.45\columnwidth]{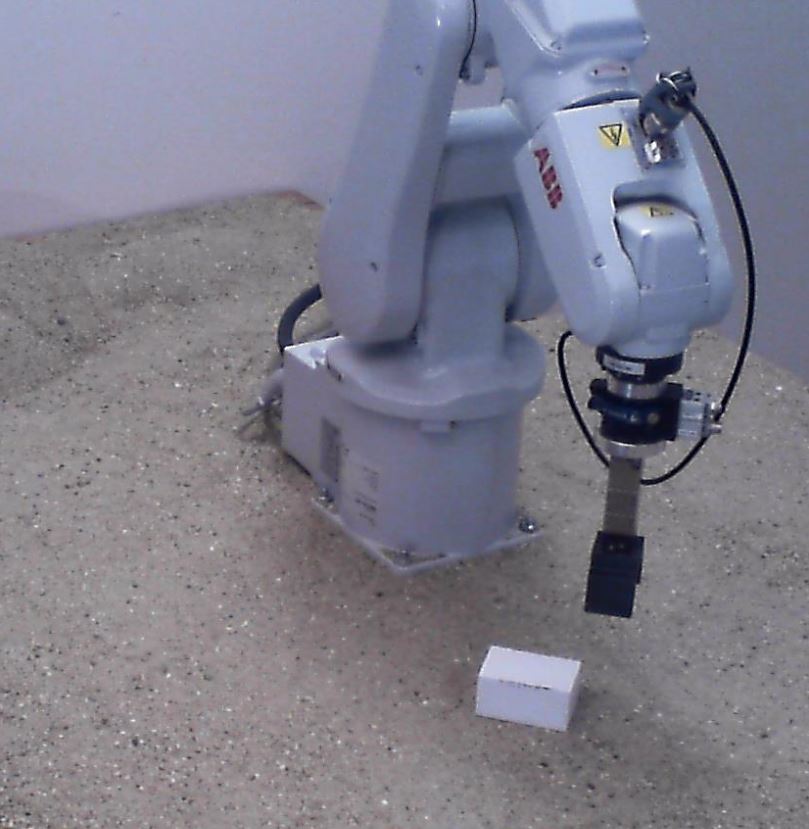}}
  \caption{We formulate relative block-robot positioning using an uncalibrated camera as a classification problem where we predict the position  $(x,y)$ of the center of a block \OLD{w.r.t.}with respect to the robot base, and the angle $\theta$ between the block and the robot main axes \textbf{\protect\subref{fig:zones}}. We show we can train a \OLD{Convolutional Neural Network (CNN)}CNN to perform this task on synthetic images, using random poses, appearances and camera parameters \textbf{\protect\subref{fig:trainimage}}, and then use it to perform very accurate relative positioning on real images \textbf{\protect\subref{fig:testimage}}}.
  \label{fig:teaser}
\end{figure}

Solving robot vision, to allow a robot to interact with the world using simple 2D images, is one of the initial\OLD{motivation} motivations of computer vision \citep{roberts1963machine}. We revisit this problem using the modern tools of deep learning\OLD{, in order to use} under assumptions that are relevant for autonomous robots in the construction industry.
In factories equipped with robots, with well-identified and repetitive tasks, cameras for robot control can be placed in both relevant and secure places, and calibrated (internally and externally) once and for all.

Now imagine using robots on a work site.  As the environment is only very weakly controlled and as the tasks can change frequently, 
there is no generic camera position that can see well all robot actions.  Even if several cameras are used to provide enough relevant views, some of these cameras would often have to be moved, either manually (e.g., hand-shifting tripods) or using cameras mounted on moving devices such as mobile observation robots or cable drones.  In these settings, taking the time to recalibrate cameras with respect to the robot after each camera movement is unrealistic, and inertial measurement units (IMUs) would not be accurate enough to recompute new usable camera poses.  On a construction site, cameras could also be hit or moved accidentally, sometimes unknowingly.  There is thus a high demand on robustness while also requiring strong accuracy for object manipulation.

An option is to directly mount cameras on the operating robots, but the practical number of cameras and the range of viewpoints is then severely reduced.  Besides, such cameras are more exposed to dirt and accidents as they are close to robot actions. \REV{Last, contrary to the relatively clean and controlled context of a factory, a construction site is a hostile environment for robot vision.  The objects suffer from substantial changes in appearance, including texture variations due to dirt and bad weather, changes of illumination due to an outdoor setting, and unruly occlusions.  There might also be shape differences between the expected 3D models and the real scene.  While this does not actually change the tasks, it makes them substantially more difficult.}

To address\OLD{this issue} these issues, we consider an alternative\OLD{set-up} setup where the robot control does not require a formal calibration step. We consider the case of several uncalibrated cameras, e.g., attached to sticks that are moved as the construction evolves, which observe a scene with no reliable landmark that would allow absolute camera\OLD{calibration} registration.  The challenge is then to recognize the different elements in the scene and position them accurately\OLD{w.r.t.} with respect to each other.

More concretely, we focus here on the problem of finely localizing and orienting a building block with respect to a robotic arm, that are both seen from external \OLD{uncalibrated}cameras \REV{with no extrinsic nor intrinsic calibration}. As argued above, this task is of major interest, but to the best of our knowledge, there are no generic method to solve it, and no standard datasets to train or test a method. In this paper, we present such a dataset, with thousands of annotated real test images covering several difficulty levels, as well as a strong baseline method learned from synthetic images.

To accomplish this task, an approach would be possible where 3D models corresponding to the robot and the building block would be first aligned on the image, then localized with respect to the camera, and finally positioned with respect to one another. However, it would be sensitive to changes in shape and appearances,\OLD{due for example to occlusions, differences between the 3D models and the real scene, and texture changes,} which are\OLD{unavoidable} pervasive on a construction site. We rather present a more direct approach which leverages recent advances in the use of\OLD{Convolutional Neural Networks (CNNs)} CNNs.\OLD{CNNs} Such networks have the advantage to be generic, they can easily be trained to be robust to many perturbations (including occlusions), and they can be applied to images extremely efficiently.

Yet, CNNs require a very large quantity of annotated training images. While it is possible to animate a robot and a building block in front of a camera to automatically create such a dataset, it requires typically thousands of hours to collect images concerning a hundred of thousands of situations \citep{PintoINCRA2016,LevineISER2017}. Given that a new training is necessary when the robot or the building blocks change, this solution is not practical. 

To make our framework generic and easily applicable to new robots and new types of blocks, without requiring the expensive and long construction of\OLD{training datasets} real-life datasets for training, we demonstrate a light and flexible method that simply takes as input the untextured 3D model of both a building block (with possible parametric variations) and a robot (with its possible joint positions), without any real image. It is based on virtual training using synthetic images only, and yet it performs well on real images.

\REV{We use a three-step approach. First, using a lower image resolution of the whole scene, we make a coarse estimation of the block pose relatively to the robot base. Second, after moving the clamp above this estimated location, we locate the clamp in the image and crop a higher-resolution image around that clamp position. Third, using this high-resolution crop, we refine the estimation of the block pose in the clamp reference system, which we can translate into the robot base reference frame as the 3D position of the clamp with respect to the base is known.}

\REV{This approach has two benefits. First, it allows to take advantage of the resolution of the input image without increasing the required memory (as a fully convolutional approach would). Second, it allows to perform a more accurate relative localization with the robot clamp close to the block, rather than with the robot base potentially far from the block.}

To summarize, our main contributions are as follows:
\begin{itemize}[topsep=0pt]
\item We argue for a new robotic task of relative localization without camera calibration, formulate it in detail, and provide a rich realistic evaluation dataset and procedure\footnote{The project page with this UnLoc dataset (Uncalibrated Relative Localization) is \url{imagine.enpc.fr/~loingvi/unloc}.}. The task involves a three-step procedure where a first coarse position estimation is refined after the robot moves towards the target.
\item We demonstrate that very accurate (millimetric) relative localization can be reached with learning techniques only despite the lack of calibration information.
\item We show that training can be performed without a single real image, paving the way for generic virtual training with arbitrary shapes and objects.
\end{itemize}


\section{Related Work}
\label{relatedWork}
There are two critical challenges in our work: first, extending localization and pose estimation to perform relative pose estimation; second performing such a task without using real images as training data. We thus review the literature on both challenges.

\subsection{Object localization and pose estimation}

Localizing an object and estimating its pose are classical computer vision problems. Indeed, the 3D understanding of the world that it provides seems to be a natural first step for image understanding and for any robotic interaction \citep{roberts1963machine}.

\paragraph{Model-based alignment methods.} In early computer vision works, it was often assumed that a 3D model of the object of interest was available \citep{roberts1963machine,lowe1987three,huttenlocher1990recognizing,mundy2006object}. The most successful approach was the use of local keypoints descriptors such as SIFTs \citep{lowe1999object}, which lead in particular to many 3D pose estimation pipelines for robotics applications, from a single or multiple images \OLD{, such as } \citep{collet2010efficient,collet2011moped}.


\paragraph{Learning-based methods.}
Learning-based approaches tend to focus more on object categories than on specific object instances. They can be roughly classified in two categories.

First, many works used manually-designed features as input to learning algorithm, and apply them to detect objects in images\OLD{(e.g., } \citep{dalal2005histograms,felzenszwalb2010object}. Deformable Part Models \citep{felzenszwalb2010object} were particularly successful and extended to predict 3D object poses \citep{glasner2011aware,fidler20123d,hejrati2012analyzing,pepik2012teaching}. 

More recently\OLD{Convolutional Neural Networks (CNNs)} CNNs \citep{lecun1989backpropagation} were shown to boost performance for 2D object detection \citep{sermanet2013overfeat,girshick2014rich,he2017mask} and object pose estimation \citep{tulsiani2015viewpoints,su2015render,wu2016single,massa2016crafting}. Even more related to our goal, such deep learning approaches were used by \citet{PintoINCRA2016} to learn where to grasp an object, and have also recently attracted much attention in the robotics community, in particular for reinforcement learning \citep{schulman2015trust,levine2016end}. 



\paragraph{\REV{Marker-based detection methods.}}

\REV{Another approach is to place one or several fiducial markers, such as ArUco \citep{GARRIDOJURADO20142280,GARRIDOJURADO2016481}, on the objects of the scene to facilitate their detection and fine localization in images.  This method was successfully used in \citet{Feng2014TowardsAR}.  It however is less flexible as it requires carefully positioning these markers on visible flat surfaces of the objects at well defined locations.  Besides, it also requires intrinsically and extrinsically calibrated cameras to locate the robot in a global reference frame (possibly putting markers on the robot too) and to relate objects in 3D.}

\paragraph{Relative positioning.}
In contrast to these works, our focus is not the localization of an object of interest in an image or with respect to the camera, but the relative positioning of one object\OLD{w.r.t.} with respect to another, the robot and the block. Rather than targeting full 6DOF pose \citep{hodavn2016evaluation}, we restrict ourselves to two position and one angle parameters, since one can reasonably assume that both the robot and the block are standing on a more or less flat surface. 
We developed a direct prediction approach, building on the success of the CNN based-approaches and following the logic of end-to-end learning, to learn to directly predict the relative pose. Both the definition of the problem of relative pose estimation and this direct approach are novelties of our work. 


\subsection{Learning from synthetic data}\label{learnfromsynth}
The success of deep learning renewed the interest in the possibilities to learn from synthetic data. Moreover, synthetic data permits to have a large amount of accurate annotations for each image. Indeed, deep learning methods typically require very large annotated datasets to be trained, but such annotations are most often expensive and difficult to obtain. However, an algorithm trained on synthetic data may not apply well to real data, because of their various differences, a problem often referred to as the domain gap between real and synthetic images. We identified\OLD{three} two main approaches to address this issue.

\paragraph{Faking realism.}
The simplest way to avoid the issues related to the differences between real and synthetic images is to try and reduce the domain gap as much as possible by generating realistic images. Generating completely realistic images requires using a high quality 3D model of a scene as well as a good illumination model, and then applying a rendering algorithm. Since high-quality 3D models are expensive and high-quality rendering is computationally expensive (typically several hours per image), little work has been done with this quality of data, and most works focus on faking realism. This can be done for example by fusing a rendered 3D object model with a texture extracted from a real image and compositing it with the background of a real scene. In particular, \citet{peng2015learning} and \citet{pepik2015holding} studied how the realism of such composite images impacted performance for object detection, \citet{su2015render} applied such a strategy to push the performance on object category pose estimation, and \citet{chen2016synthesizing} on human pose estimation. More recently, several papers have used game engine to generate training data, for example for semantic segmentation \citep{richter2016playing,ros2016synthia,shafaei2016play}.

\paragraph{Domain adaptation.}
The problem of training a model on synthetic images and applying it to real images can be seen as an instance of the generic domain adaptation problem, for which many approaches have been developed.\OLD{A detailed presentation of these methods is beyond the scope of this paper and a recent overview can be found in .} 
One approach is to explicitly design a loss or an architecture to perform domain adaptation. Such methods have been used with synthetic data to perform object classification \citep{peng2017synthetic}, detection \citep{sun2014virtual,vazquez2014virtual} and, closer to our task, 3D model alignment in 2D images \citep{massa2016deep}.

\begin{figure}[t]
  \centering
  \subfloat[Synthetic training image and real testing image from \citet{tobin2017domain}.]{\label{fig:comp}\includegraphics[height=0.44\columnwidth]{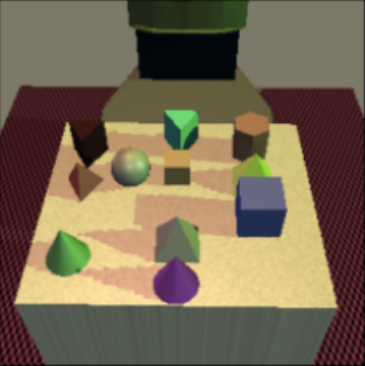}~\includegraphics[height=0.44\columnwidth]{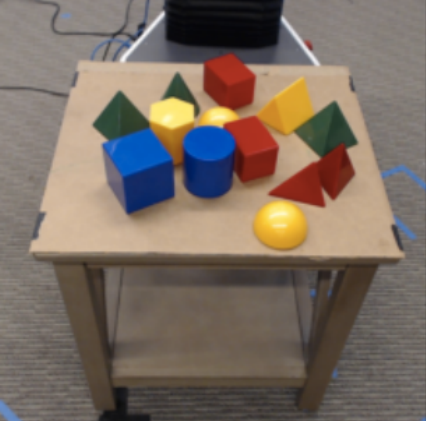}
  }
\newline
  \subfloat[Our synthetic training image and real testing image.]{\label{fig:ours}\includegraphics[width=0.44\columnwidth]{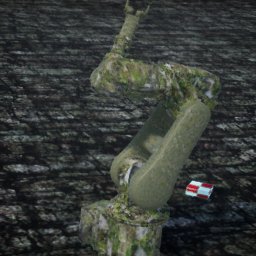}~\includegraphics[width=0.44\columnwidth]{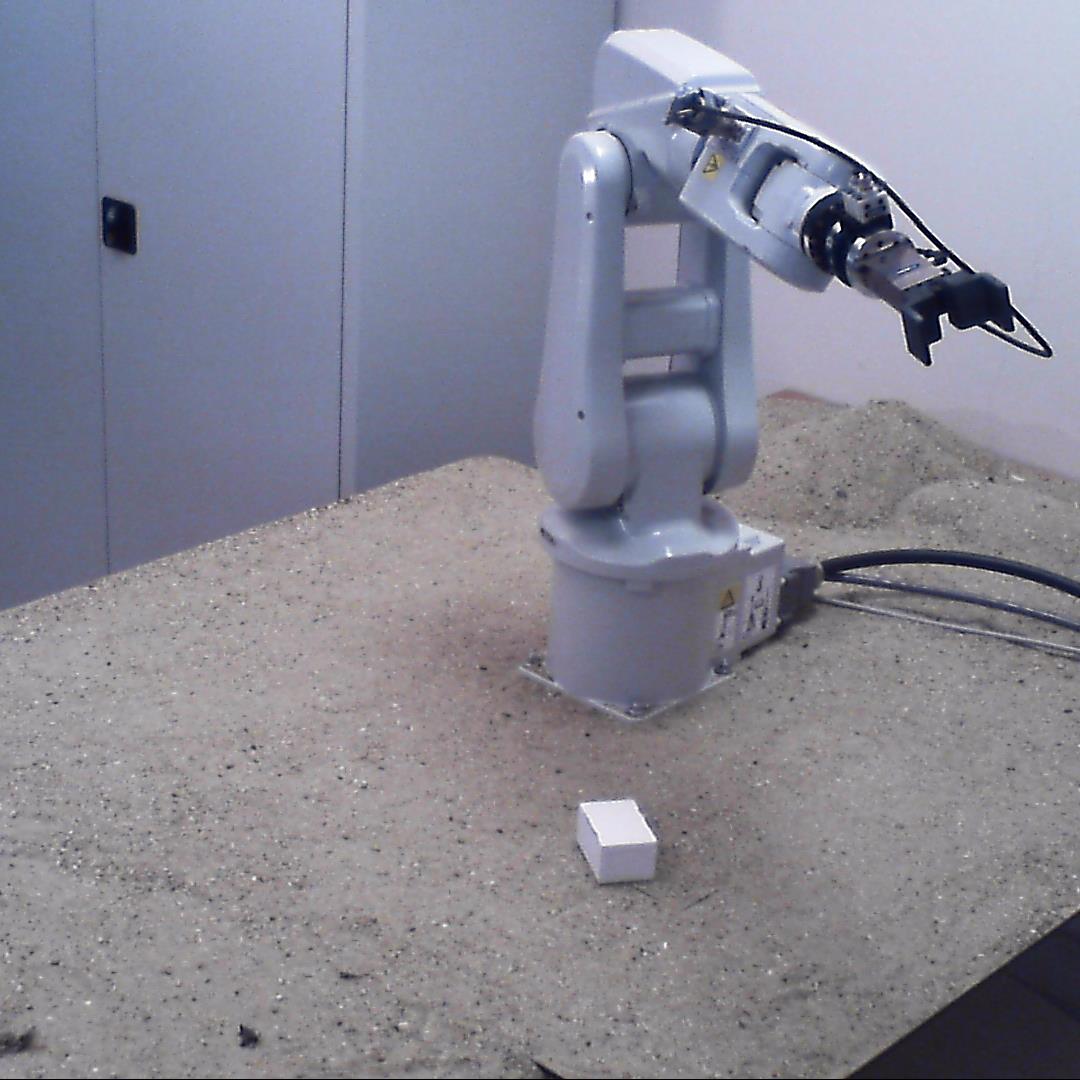}
  }
  \caption{Qualitative comparison of the recent work of \citet{tobin2017domain} with our approach. While \citet{tobin2017domain} focuses on the differentiation between objects on a fixed size table, we locate a block with respect to a robot in an unknown configuration and in an unknown environment, observed by a camera in an unknown pose}
  \label{fig:relatedw}
\end{figure}

\OLD{The last}Another approach for learning robust models with synthetic data is to generate training examples with a very high variety to encourage the models to learn invariances. The first application of such an approach with\OLD{convolutional neural networks} CNNs is probably \citet{dosovitskiy2015flownet}, which leverages highly-modified and unrealistic composite images between real images and rendered 3D chairs \citep{aubry2014seeing} to learn to predict optical flow between real images.
More recently \citet{sadeghi2016cad} demonstrated how to use a high variety of non-realistic rendered views of indoor scenes to learn how to fly a quadcopter.


Independently of our work,\OLD{Tobin et al.} \citet{tobin2017domain} recently proposed a similar strategy, that they call domain randomization, for localizing a block and grasping it with a robot. There are however several strong differences between this work and ours. In particular,\OLD{Tobin et al.} \citet{tobin2017domain} focus on the differentiation between different shapes of objects (triangular, hexagonal, rectangular) and coarse localization on a table with fixed characteristics, i.e., always seen from a similar viewpoint, with the same orientation. Thus, \citet{tobin2017domain} essentially requires localizing the table and objects in 2D. On the contrary we do not impose the presence of a table with fixed characteristics and seen from a given angle; we localize the objects directly with respect to a robot at unknown positions. These differences in our data are illustrated on Figure~\ref{fig:relatedw}. Furthermore, our approach allows to localize a block with an average accuracy of \OLD{3.5\,mm and 1 degree} \OLD{2.5\,mm and 0.7 degree}\REVV{2.6\,mm and 0.7\textdegree}, while \citet{tobin2017domain} provide an average accuracy around 15\,mm, without orientation estimation. 


\section{Task Abstraction\OLD{ and Evaluation}}
\label{sec:method}





The locating-for-grasping problem is a robotic task that is difficult to assess. In this section, we first explain how we express it as a\OLD{two} three-step procedure corresponding to\OLD{two} three pure computer-vision subtasks, which enables reproducible quantitative evaluation. We then formalize these subtasks. \OLD{, and finally describe an evaluation dataset.}


\subsection{Task overview}

The task we consider is as follows. The goal is to be able to grasp an unknown cuboid block with a robotic arm using only images from \REV{intrinsically and extrinsically} uncalibrated cameras. The block lies on a mostly planar ground; its position, orientation and sizes are unknown.  The robot and the block are in the field of view of several cameras, whose actual position and orientation are also unknown; as a result, the block can be fully or partially occluded by the robot on some camera views.

For grasping to succeed, the position and orientation of the cuboid \emph{with respect to the robot} must be known accurately. Please note however that only this \emph{relative} position and orientation between the robot and the block is required; the actual camera poses are not needed.

The only configuration information that is available for this task is an \emph{untextured} 3D model of the robot with its joint axes, and a range of possible sizes for the cuboid. In particular, the aspects of both the robot and the block are totally unknown.

As it remains difficult to accurately locate an object with respect to a robot, we actually define the following\OLD{two} subtasks, which correspond to a\OLD{two} three-step locating procedure:
\begin{enumerate}[topsep=3pt,itemsep=3pt]
\item \REV{\emph{Coarse relative localization subtask.}} We first consider the very general case where the robot and the block are at random positions and orientations, and the robot joints are also randomly set. The cameras, which are random too, only provide overviews of the scene.\OLD{that are used} \REV{The subtask here is to (coarsely)\OLD{locate} estimate the pose of the block with respect to the robot}.

\item \REV{\emph{Tool localization subtask.}} After the block position is thus estimated, although possibly with\OLD{a lack of} moderate accuracy and confidence, we assume the robot clamp is moved on top of that coarse predicted location.  In this setting, the robot and the block remain at random positions and orientations, but the clamp is located at a random position close to the block, oriented towards the ground and ready to grasp. \REV{Now the second subtask is to detect the clamp in the picture, allowing camera close-ups to later perform a finer pose estimation.}

\item \REV{\emph{Fine relative localization subtask.}} Last, using camera close-ups (actually crops of overviews centered on the zone of interest),\OLD{can then be used} \REV{the third subtask is to finely estimate the block location and orientation with respect to the clamp}, hence with respect to the robot, thus enabling the actual grasp.
\end{enumerate}

\REV{Note that in this paper, we focus on the position estimation in the horizontal plane where the robot rests and assume the block lies flat on this plane with little or no variation of height or tilt.  It corresponds to the realistic assumption that we are working on an approximately planar surface. Adapting the framework to deal with significant block tilts and height variations is future work.}

We describe in Section~\ref{sec:realdata}\OLD{below} a dataset to assess methods that try to address these\OLD{two} three subtasks. It is made of real images of a robot and a block, together with accurate annotations of the relative position and orientation of the block with respect to the robot. This allows to quantitatively evaluate the success of a method on the task.

This composite task actually abstracts a more general task of image-based object-robot relative\OLD{location} positioning, where robot motions are arbitrary.  However, as such a visual servoing is a dynamic process that involves real devices, it is practically impossible to reproduce and hence, it is not possible to compare two given methods. Our\OLD{two} three-step locating task for grasping is a discrete formulation of this dynamic process, with a predefined plausible intermediate move. The interest is that it is static and deterministic; competing methods can thus be assessed and compared quantitatively and meaningfully.

Our experiments below also show that such a\OLD{two} three-step procedure makes sense for robot control as indeed more accurate position and orientation information can be obtained thanks to the\OLD{additional, second} refinement subtask.
\REV{Additionally, we consider two task variants: single-view and multi-view.
In fact, as the block can sometimes be occluded in a single view, using multiple views provides more robustness because then a block has little or no chances to be hidden in all views. The multi-view setting also offers a greater pose accuracy by exploiting information from each views.
}

\subsection{Robot and reference frames}

\begin{figure}[t]
\centering
\includegraphics[width=0.65\columnwidth]{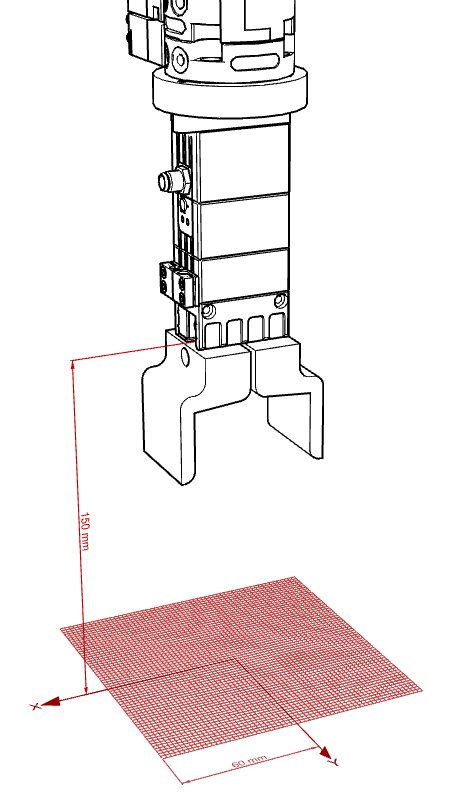}
\caption{Reference frame for fine location estimation.  Note the small asymmetry between the left and right sides of the clamp (plate and cable connector).  In our experiments, to estimate block position, we use a 12\,cm grid with 2\,mm steps (best seen with electronic version)}
\label{fig:zoom}       
\end{figure}

We need to define a reference frame to formalize the positioning subtasks. As the blocks are symmetric (invariant to a 180\textdegree\ rotation in the horizontal plane), and as the problem is mostly robot-centric, we choose to associate the reference frame to the robot.

For frame axes to be defined meaningfully, the robot has to have asymmetries.  Rather than imposing markers on the robot, which are not robust, especially in case of a construction site with dirt and possibly bad weather, we choose to rely only on the asymmetries of the robot intrinsic shape.

Concretely, in the dataset we created for this task (see Sect.~\ref{sec:realdata}), we used a robot (an IRB120 from ABB company) with the following shape features:
\begin{itemize}[topsep=3pt,itemsep=3pt]
\item The base of\OLD{our} the robot is mostly cylindrical. However, it is fixed on the experiment table with nuts and bolts using a mostly-square plate whose orientation can be used to define axes $X$ and $Y$, up to swapping. More importantly, wires are connected to the robot via a square box attached to the base cylinder, which we use to define the direction of the $X$ axis; the $Y$ axis is then defined as orthogonal and oriented by convention.  The block relative orientation $\theta$ is the angle between the largest block dimension (on the ground plane) and the $X$ axis.  As our cuboid blocks here are symmetric, this angle is defined only between 0 and 180 degrees. This frame is illustrated on Figure~\ref{fig:zones}. It is used for the coarse location subtask.
\item Just above the clamp, a bolt and a small square cover make the side of the grip support asymmetric, as shown on Figure~\ref{fig:zoom}. As above, it allows us to define two axes $X$ and $Y$, and the angle $\theta$ of the main block axis with respect to axis~$X$. This frame is relative to the clamp rather than to the robot base, but it can immediately be related to the robot frame as the robot joints and clamp position are known. Note that this second frame only makes sense when the clamp is oriented vertically towards the ground. It is used for the fine location subtask.
\end{itemize}
Similar shape asymmetries can easily be added to any existing robot if they do not already exist. 



\section{Solving the Task with\OLD{a} CNNs and Synthetic Images}
\label{sec:ourmethod}

Recovering the exact position of an object (robot, things to grasp, obstacles, etc.) from a single image given only a 3D model is a difficult problem, especially when the object texture is unknown or unreliable, and when its deformations (robot articulations, variants of object size and shape) make it vary drastically. Yet, given several views, the cameras could be calibrated and 3D reconstruction could be performed, enabling the problem to be solved after 3D model alignment. However, multi-view calibration and reconstruction can be challenging problems, requiring a significant processing time, 
and aligning 3D models to a point cloud is not trivial either. 

\OLD{Thus, rather than trying to define a multiple-step pipeline, with potential errors introduced at each complex stage}Instead, we propose to tackle the problem directly and train a CNN to localize a block robustly with respect to the robot, given a single image. \REV{In case images from several cameras are available, location estimates can be aggregated to improve the general accuracy.} Moreover, we only use synthetic training images, to avoid the costly collection and annotation of real data. 

\REV{More precisely, we train three CNNs, one for each subtask. The first network estimates a coarse location of the block relatively to the robot base in a single image.  After the robot clamp has been moved above that rough location, the second network locates the clamp in a view of the new scene configuration.  Last, using an image crop around the estimated clamp position, the third network estimates a fine block pose with respect to the clamp, hence to the robot base.}

\REV{While there are several steps in this approach, it differs from a pipeline with camera calibration, 3D reconstruction and model alignment in that the errors do not accumulate.  Indeed, assuming the first and second networks are robust enough to picture a close-up of the clamp with the block,\OLD{then} the final third network can make a fine estimation that does not depend on the accuracy of the previous subtasks, including crop centering.}

\subsection{Relative pose estimation with\OLD{CNNs} a CNN}
\label{sec:classification}

\begin{figure*}[t]
 \centering
  \subfloat[Sample of training data for coarse pose estimation (of block relatively to robot base)]{\label{fig:synthcoarse}
   \includegraphics[width=0.2\columnwidth]{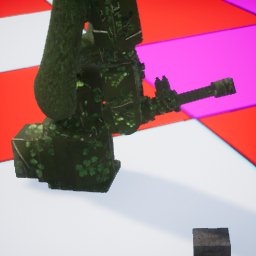} 
    \includegraphics[width=0.2\columnwidth]{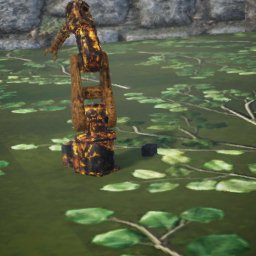} 
      \includegraphics[width=0.2\columnwidth]{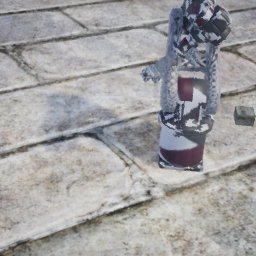} 
   \includegraphics[width=0.2\columnwidth]{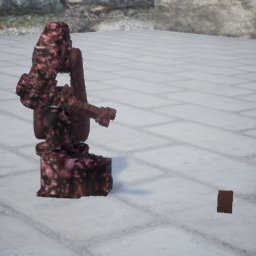} 
     \includegraphics[width=0.2\columnwidth]{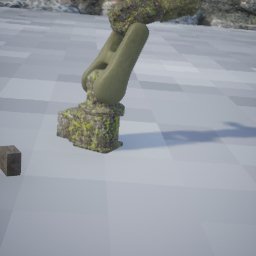} 
      \includegraphics[width=0.2\columnwidth]{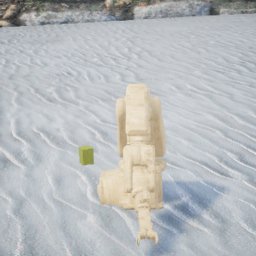}
      \includegraphics[width=0.2\columnwidth]{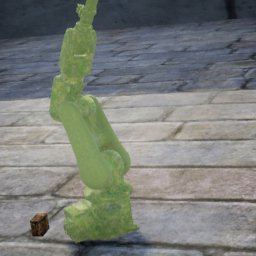} 
       \includegraphics[width=0.2\columnwidth]{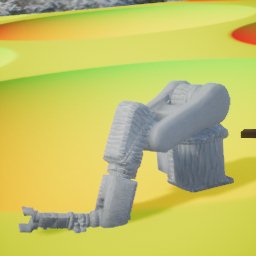} 
        \includegraphics[width=0.2\columnwidth]{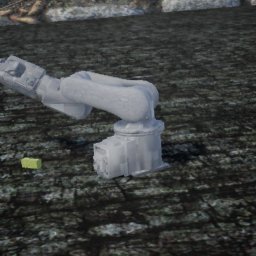} 
         \includegraphics[width=0.2\columnwidth]{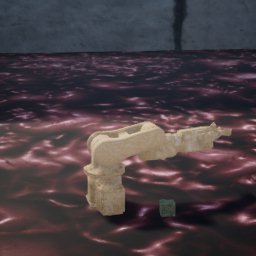}
         }
  \\
  \subfloat[Sample of training data for clamp 2D detection]{\label{fig:synthclamp}
      \includegraphics[width=0.2\columnwidth]{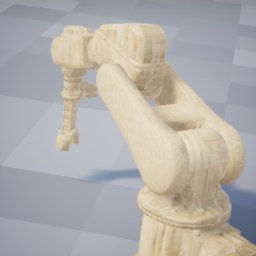} 
       \includegraphics[width=0.2\columnwidth]{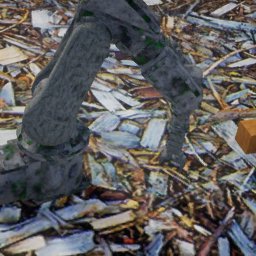} 
        \includegraphics[width=0.2\columnwidth]{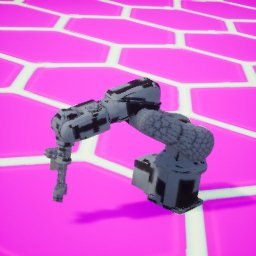} 
         \includegraphics[width=0.2\columnwidth]{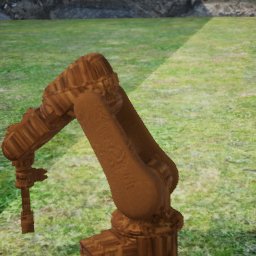} 
      \includegraphics[width=0.2\columnwidth]{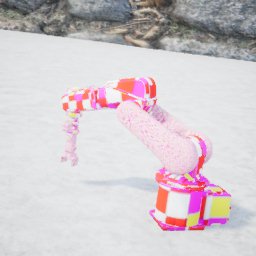} 
       \includegraphics[width=0.2\columnwidth]{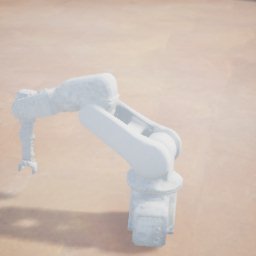} 
        \includegraphics[width=0.2\columnwidth]{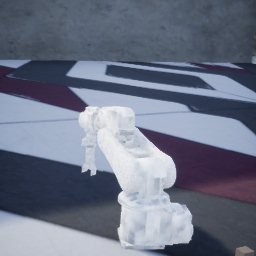} 
      \includegraphics[width=0.2\columnwidth]{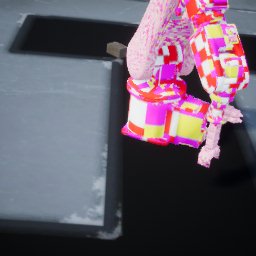} 
       \includegraphics[width=0.2\columnwidth]{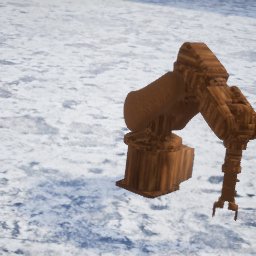} 
        \includegraphics[width=0.2\columnwidth]{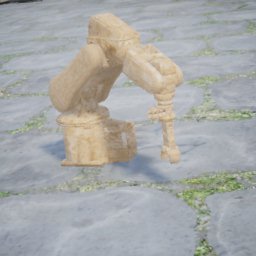} 
      }
 \\
  \subfloat[Sample of training data for fine pose estimation (of block relatively to clamp)]{\label{fig:synthfine}
      \includegraphics[width=0.2\columnwidth]   {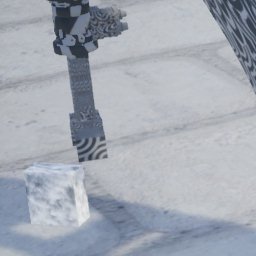} 
       \includegraphics[width=0.2\columnwidth]{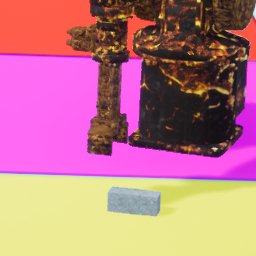} 
        \includegraphics[width=0.2\columnwidth]{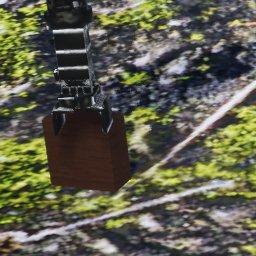} 
         \includegraphics[width=0.2\columnwidth]{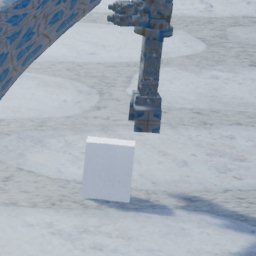} 
      \includegraphics[width=0.2\columnwidth]{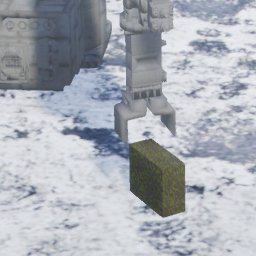} 
       \includegraphics[width=0.2\columnwidth]{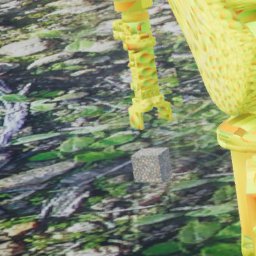} 
        \includegraphics[width=0.2\columnwidth]{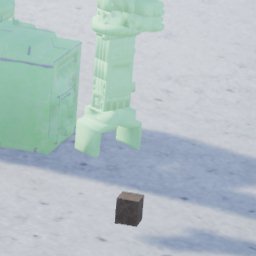} 
      \includegraphics[width=0.2\columnwidth]{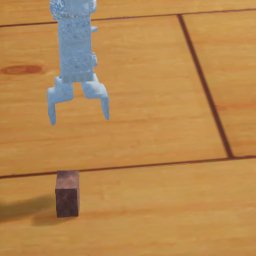} 
       \includegraphics[width=0.2\columnwidth]{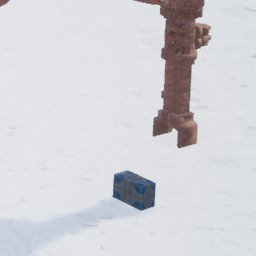} 
        \includegraphics[width=0.2\columnwidth]{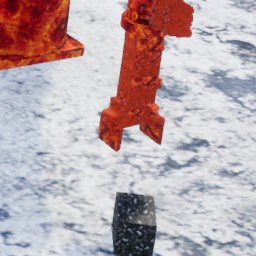} 
      }
 \caption{Our synthetic training dataset covers a large variety of random viewpoints, joint positions, object poses and sizes, as well as texture (for the robot, the block and the environment), making the trained model robust to all of these parameters}
  \label{fig:synth}
\end{figure*}

\paragraph{Classification approach.}
Since the $(x,y)$ position as well as the angle~$\theta$ are continuous quantities, it would be natural to formulate their estimation as a regression problem. However, such a formulation is not well suited to represent ambiguities. For example, since the arm of the robot is almost symmetric, it would be natural for our system to hesitate between two symmetric positions during the fine estimation step. In a regression set up, there is no natural way to handle such ambiguities. On the contrary, if we were to predict probabilities for each position of the block, a multi-modal distribution could be predicted in an ambiguous case. We thus discretize the space of possible poses and predict probabilities for each bin, formulating the problem as a classification problem. Such a formulation has been shown to be more effective in similar problems, such as keypoint position estimation and orientation estimation \citep{tulsiani2015viewpoints,su2015render,massa2016crafting}.  Concretely, given the level of accuracy we target (see Section~\ref{sec:metrics}), our coarse pose estimation considers square bins of size 5\,mm, and our fine pose estimation uses bins of size 2\,mm.

\paragraph{Network architecture and joint prediction.}

We solve the classification problem we have just defined by training a CNN. More precisely, we use a ResNet-18 network, \REV{trained from scratch} and with a standard cross-entropy loss, which has shown good performance while remaining relatively light. \REV{We used a batch size of 128, a weight decay of $10^{-4}$ and a momentum of 0.9. We trained our model with an initial learning rate of $10^{-2}$ until convergence (250 iterations for the coarse estimation, 11 iterations for the tool detection, and 186 iterations for the fine estimation), then a learning rate of $5.10^{-3}$ for 15 iterations (coarse and fine)}. We did not explore in detail the influence of the size of the network, but one can reasonably expect a small performance gain by optimizing it, i.e., using a deeper or wider network while keeping it small enough to avoid overfitting.

However, the definition of the form of the network output is not trivial. Simply defining one class per bin would lead to too many classes. For example, there are more than 17,500 attainable 2D positions bins for the coarse estimation (3,600 for the fine estimation), each of which could be broken into 36 orientation bins (90 for the fine estimation). This would not only make the network memory usage much larger, but also require much more training data. Indeed, we show in our early experiments (see Section~\ref{sec:netarchi}) that, despite the intuitive information sharing that could be used to predict the different classes, 80 images per localization class (for a total of more than 280k training images) are not enough to avoid overfitting.

The simplest alternative would be to simply train three independent networks to predict $x$, $y$ and the angle $\theta$. However, we show that it is better to train a multi-task network that predicts probabilities for different 1D bins for $x$, $y$ and $\theta$ independently, but computes a single representation in all but the last layer. Interestingly and contrary to the network prediction with one probability per 2D position, this network does not overfit dramatically.



\paragraph{Multi-view prediction.} Performing classification instead of regression also has the advantage that it gives a natural way to merge prediction from several views. Indeed, the output of the network can be interpreted as log probabilities. If one considers the information from the different views to be independent, the position probability knowing all the images is simply the product of the individual probabilities. One can thus simply predict the maximum of the sum of the outputs of the network applied to the different images.

In fact, as estimation from a single view can be ambiguous, for example because the exact size of the block is unknown or because the block is occluded by the robot, merging the predictions from several views is crucial to the success of our approach, as shown in Section~\ref{sec:resultsaggregation}.  \REV{Note that this approach not only scales well with the number of cameras but also, conversely, allows a smooth degradation (up to a single view) in case some cameras are moved or turned away from the scene, or become inoperative.}

More complicated strategies could of course be used, such as training a network to directly take several views as input, or training a recurrent neural network that would "see" the different views one by one.

\label{sec:scenario}

\subsection{\OLD{Generation of}Synthetic training dataset\OLD{base}}
\label{sec:database}

The creation of a large training dataset of annotated real photographs with variations representative of those encountered in actual test scenarios is difficult and time consuming, especially when considering the variability of uncontrolled environments, e.g., related to dirt or outdoor illumination. Instead, we generated\OLD{a} three synthetic sets of rendered images \REV{together with ground-truth pose information,} to use as training and validation sets for each subtask.
Rather than spend a lot of processing time to generate photorealistic images or use some form of domain adaptation (see Section~\ref{learnfromsynth}), we chose to apply a strategy of massive image generation (hundreds of thousands of non-photorealistic images) with massive randomization targeted at the specific properties we want to enforce.  In this synthetic dataset, as illustrated on Figure~\ref{fig:synth}, the robot is placed on a flat floor and a cuboid block is laid flat nearby, in the following configurations:
\REV{
 \begin{enumerate}
  \item robot and block in random poses, for coarse estimation,
  \item robot with clamp in random vertical pose, for 2D tool detection,
  \item close-up on vertical clamp with random block nearby, for fine estimation.
 \end{enumerate}}
To encourage our model to generalize as much as possible, 
we introduced the following randomization in the generation of synthetic images:
\REV{%
\begin{itemize}[topsep=3pt]
\item robot base orientation and position, and joint angles,
\item (cuboid) block dimensions, orientation and position,
\item textures for the floor, robot and block,
\item camera center, target, rotation and focal length.
\end{itemize}
Details on this synthetic dataset are given in Appendix~\ref{sec:syntheticdataset}.}

\section{Experimental Results and Analysis}
\label{sec:results}
In this section, we\OLD{use new evaluation datasets to} describe an evaluation dataset and use it to analyze in detail the performance of our direct learning-based approach, including the different design choices.

\subsection{Evaluation dataset}
\label{sec:realdata}

As far as we know, there exists no dataset concerning the task of high-precision object-robot relative localization. We created such a dataset, with real images, for the composite task described in Section~\ref{sec:method}.  Note that it is only a test (evaluation) dataset, not a training dataset.

\REV{The dataset is divided in three parts, according to each subtask:
\begin{enumerate}[label=(\arabic*),topsep=3pt,itemsep=3pt]
\item given an image of the robot and a block, find their (coarse) relative pose in the support plane,
\item given an image of the robot with a vertical clamp pointing downwards, find the clamp location in the image,
\item given a zoomed image of a vertical clamp pointing downwards and a block, find their (fine) relative pose in the support plane.
\end{enumerate}
The relative block poses are consistent across the three parts of the dataset: for each random block pose, we picture:
\begin{enumerate}[label=(\roman*),topsep=3pt,itemsep=3pt]
\item a long shot of the scene where the robot joints are set at random angles,\label{robotrandom}
\item a long shot of the scene where the clamp is moved near the block and set vertical and pointing downwards,\label{clampnearblockfar}
\item a crop of that same image, more or less centered on the clamp and large enough to show the block as well.\label{clampnearblockclose}
\end{enumerate}

\begin{figure}
 \centering
 \includegraphics[width=0.9\columnwidth]{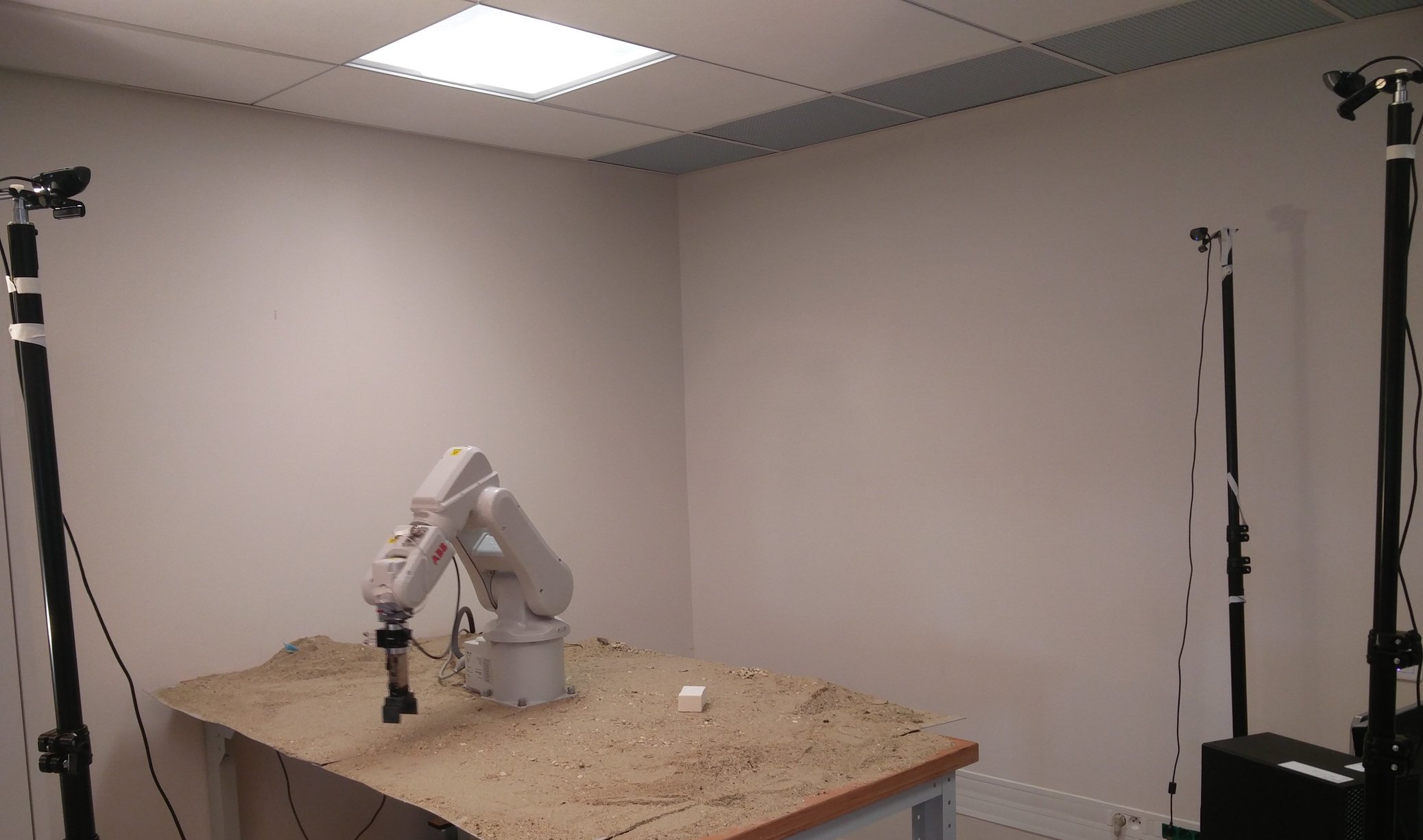}
\caption{Acquisition of real images to create our test dataset, with 3 cameras at arbitrary and varying positions}
\label{fig:acquisition}       
\end{figure}

\begin{figure}[t]
 \centering
 \includegraphics[width=0.43\columnwidth]{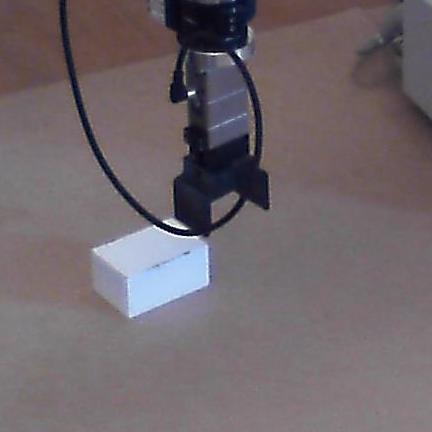} 
  \includegraphics[width=0.43\columnwidth]{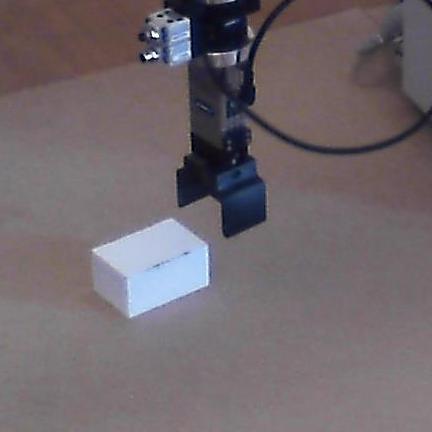}
 \caption{Example of the clamp roughly located above the block, and the two clamp positions (P1 and P2) used to possibly improve the robustness of block pose estimation}
 \label{fig:gripor}
\end{figure}

To study the benefits of having multiple views of the scene, each configuration is actually seen and recorded from 3 cameras.  Moreover, as the clamp we used only has a small asymmetry on one side (see Figure~\ref{fig:zoom}), leading to a possible direction ambiguity when estimating the clamp frame, we recorded for each configuration of type \ref{clampnearblockfar}-\ref{clampnearblockclose} an additional position where the clamp is turned vertically 90\textdegree (see Figure~\ref{fig:gripor}.

\begin{figure}[t]
  \centering
  \subfloat['lab' dataset.]{\label{fig:labdataset}\includegraphics[height=0.4\columnwidth]{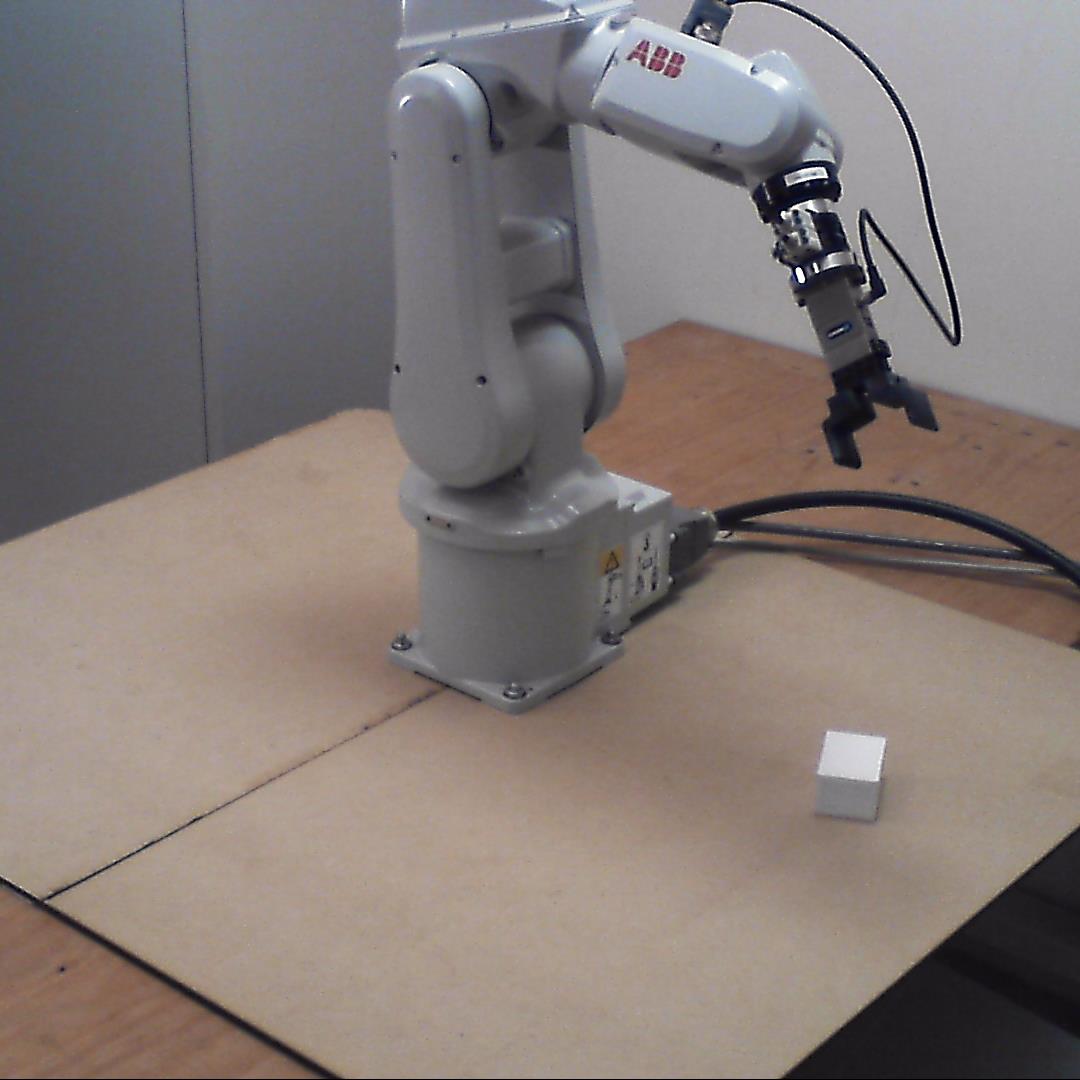}~~~
  \includegraphics[height=0.4\columnwidth]{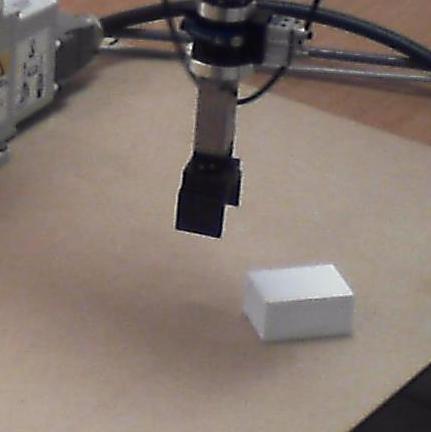}
  }\\
  \subfloat['field' dataset.]{\label{fig:fielddataset}\includegraphics[width=0.4\columnwidth]{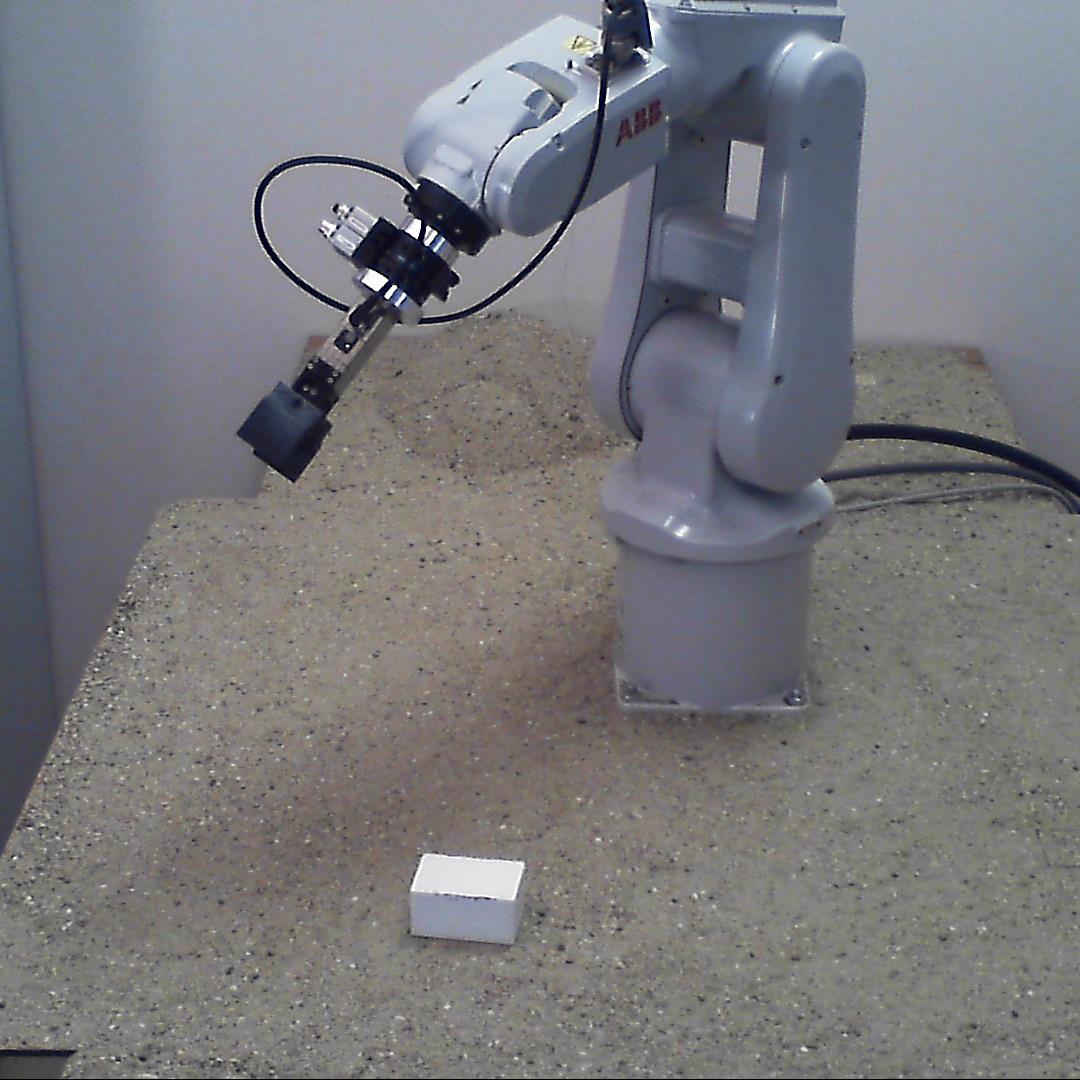}~~~
  \includegraphics[width=0.4\columnwidth]{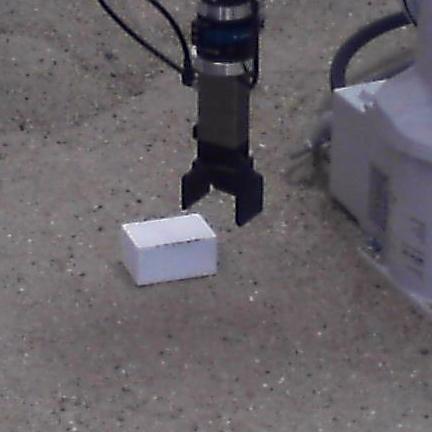}
  }\\
  \subfloat['adv' dataset.]{\label{fig:advdataset}\includegraphics[width=0.4\columnwidth]{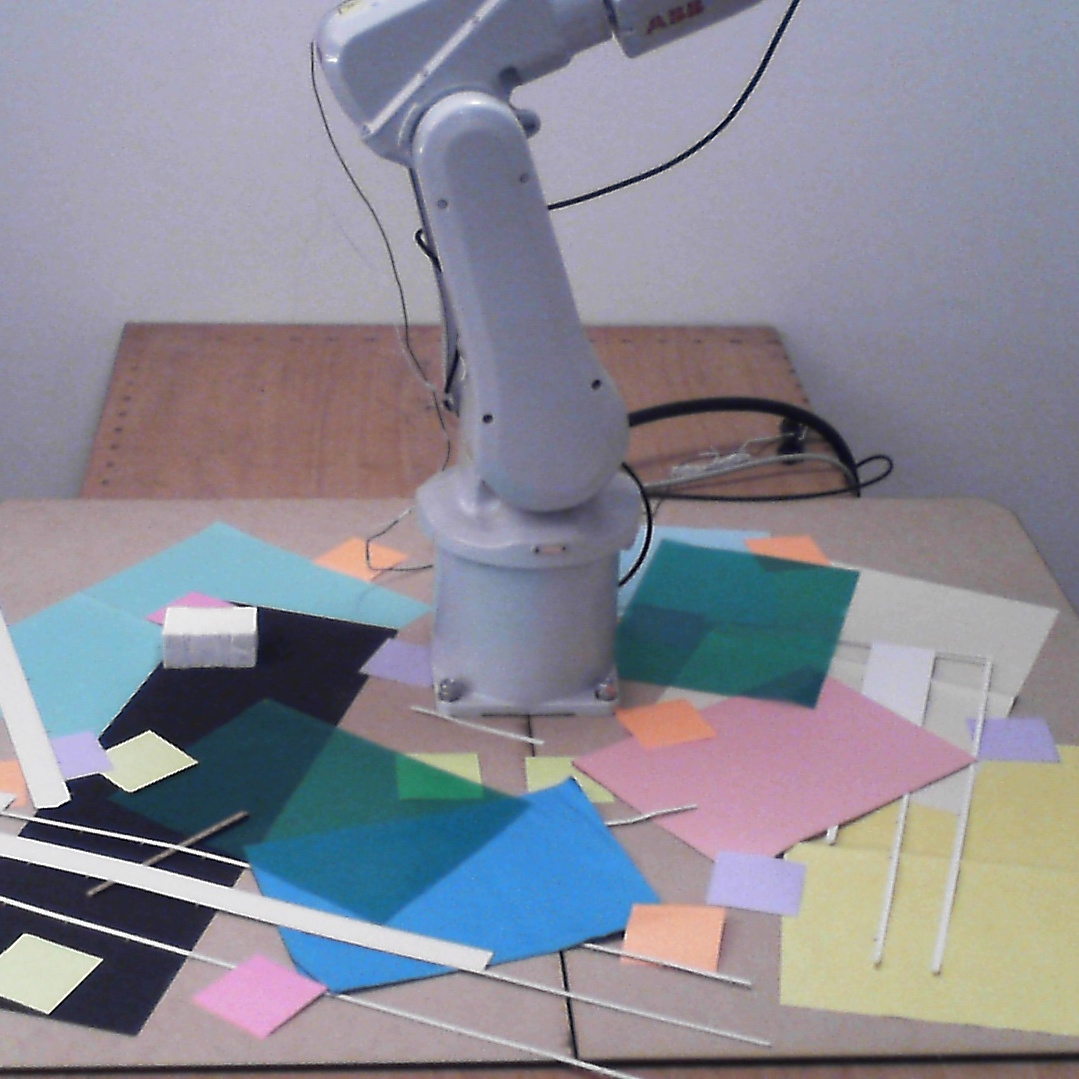}~~~
  \includegraphics[width=0.4\columnwidth]{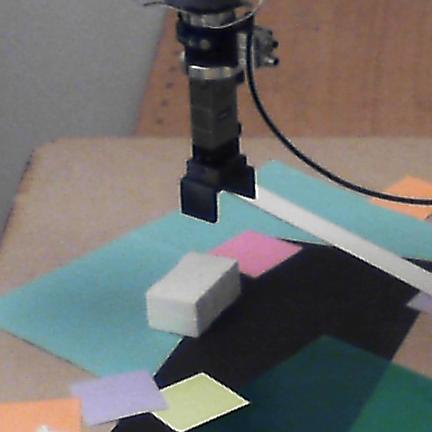}
  }
  \caption{Example images from the three variants of our evaluation dataset, for the coarse localization (left) and fine localization tasks (right): clean laboratory conditions~\textbf{\protect\subref{fig:labdataset}}, more field-like conditions with gravels~\textbf{\protect\subref{fig:fielddataset}}, and adverse conditions with distractors~\textbf{\protect\subref{fig:advdataset}}}
  \label{fig:datasets}
\end{figure}

Additionally,} we\OLD{actually} made three variants of each of these (sub) datasets, corresponding to different environment difficulties, as illustrated on Figure~\ref{fig:datasets}:
\begin{enumerate}[label=(\alph*),topsep=3pt,itemsep=3pt]
\item a dataset in laboratory condition ('lab'), where the robot and the block are on a flat table with no particular distractor or texture,

\item a dataset in more realistic condition ('field'), where the table is covered with dirt, sand and gravels, making the flat surface uneven, with blocks thus lying not perfectly flat, and making the appearance closer to what could be expected on a real construction site,

\item a dataset in adverse condition ('adv'), where the table is covered with pieces of paper that act as distractors because they can easily be confused with cuboid blocks.
\end{enumerate}
\REV{All together, the whole dataset covers about 1,300 poses (576 for 'lab', 639 for 'adv', and 114 for 'field'), seen from 3~cameras, with 1~clamp position for configuration~\ref{robotrandom} and 2~clamp positions for configuration \ref{clampnearblockfar}-\ref{clampnearblockclose}, yielding about 11,700 annotated images. More details on this evaluation dataset are given in Appendix~\ref{sec:evaluationdtataset}.}

In the following, to study different properties or design choices of our approach, we report results on the 'lab' dataset only, except otherwise mentioned.

\subsection{\REV{Evaluation metrics}}\label{sec:metrics}

To assess pose estimation results $x$, $y$, $\theta$, several measures can be considered. The most natural one is to measure the \emph{accuracy}, computing the average and standard deviation of the estimation errors $e_x$, $e_y$, $e_\theta$. While this provides figures that can be easily interpreted, it is not a useful measure of the capacity of a method to provide an accurate-enough prediction as it does not take into account a maximum possible error.  We thus also present \emph{success} measures in terms of the rate of estimation errors $e_x$, $e_y$, $e_\theta$ that are below given thresholds, i.e., that are accurate enough for block grasping to be successful.

\REV{Concretely, for the fine estimation network, we consider a square range of size 12\,cm, as illustrated on Figure~\ref{fig:zoom}.  Fine estimation thus makes sense if the prediction error of the coarse estimation is below 6\,cm. Besides, for a grasp to be successful given the opening of the clamp we used in the dataset, the required accuracy is 5\,mm for $x$, 5\,mm for $y$ and 2\textdegree\ for $\theta$.} 
Therefore, for the coarse estimation, we report the percentage of prediction errors below 6\,cm for the block location $x$ and $y$; we also consider an error bound of 10\textdegree\ for the block orientation~$\theta$. 
For the fine estimation, the error bounds are 5\,mm for block location, and 2\textdegree\ for block orientation.

\REV{Regarding the clamp location estimation in images (second, intermediate subtask), we did not create a ground truth. (We would have had to run a full camera and robot calibration.)  Instead, for our experiments, we manually checked all predictions for localizing the center of the clamp, and counted cases where the prediction was outside the bounding box of the clamp in the image.}


\subsection{Network architecture}\label{sec:netarchi}

\REV{As mentioned in Section~\ref{sec:classification}, we chose to predict independently $x$, $y$ and $\theta$ using a single network which computes a single representation in all but the last layer. To validate our choice, we evaluated the accuracy in the prediction of the $x$ coordinate for our \OLD{fine}\REVV{coarse} estimation setup 
with three different networks. These networks predict the block pose with three different approaches, either:}

\begin{table}
\caption{\OLD{Performance for the}Success rate of \OLD{fine}\REVV{coarse} pose estimation\OLD{with} for three different network architectures in a \OLD{3}\REVV{1}-camera setting for the 'lab' dataset, measured as the percentage of estimation errors for $x$ below 60\,mm. 
}
\label{tab:archi}
\centering
\vspace{-2mm}
\begin{tabular}{l|@{~}c@{~}|@{~}c@{~}|@{~}c}
\noalign{\smallskip}\hline\noalign{\smallskip}
\OLD{network} Architecture  & 1D $x$ bins 
& 2D $(x,y)$ bins & 1D $x$, $y$ and \OLD{angle}$\theta$ bins    \\
\noalign{\smallskip}\hline\noalign{\smallskip}
\% ($e_x$ $\leq$ \REVV{60}\,mm)   &  $\OLD{64.8}\REVV{97.0}$ & $\OLD{11.8}\REVV{6.1}$ & $\OLD{72.8}\REVV{98.7}$  \\
\noalign{\smallskip}\hline
\end{tabular}
\end{table}

\OLD{\REV{To validate our choice of predicting separately the different dimensions with a single classification network we evaluated the accuracy in the prediction of the $x$ coordinate for our fine estimation setup, aggregating the information of one view from each of the three cameras as explained in Section~\ref{sec:classification}, with three different networks and report the results in Table~\ref{tab:archi}. This networks predict the block pose with three different approach, either}:}
\begin{itemize}[topsep=0pt]
\item \REV{along the $x$ axis only, with bins of width \OLD{2}\REVV{5}\,mm,} \OLD{one network predicting only probabilities for each 2\,mm bin of $x$,}
\item \REV{with $(x,y)$ square bins of size \OLD{2}\REVV{5}$\,\times\,$\OLD{2}\REVV{5}\,mm,} \OLD{one network predicting probabilities for each 2$\,\times\,$2\,mm square in the plane,}
\item \REV{along both axes $x$ and $y$ separately, with bins of width \OLD{2}\REVV{5}\,mm, and for orientation $\theta$, with bins of size \OLD{1}\REVV{5}\textdegree.} \OLD{one network predicting probabilities for each 2\,mm bin of $x$ and $y$ and 1\textdegree\ bins of orientation.}
\end{itemize}
The results are reported in Table~\ref{tab:archi}. They clearly show that the network predicting\OLD{probabilities} location for each \OLD{2}\REVV{5}$\,\times\,$\OLD{2}\REVV{5}\,mm square is failing, even if it could in theory represent more complex ambiguities in the position estimation. Analyzing the performance of training and validation on the synthetic dataset shows that it actually overfits. On the contrary, the network trained for\OLD{1D} estimating $x$ only performs well, apparently being able to generalize across several $y$ positions. The network predicting separately $x$, $y$ and the angle $\theta$ performs even better, showing a clear benefit to the joint training. We use the latter architecture for all the following experiments. More technical details are given in Appendix~\ref{sec:netdetails}.

\subsection{Learning on synthetic data, testing on real}

Our networks are trained on synthetic data only.  One of the first question is how they behave on real images. \REV{In this section, we compare the performance of our fine pose estimation network both on synthetic and real validation data.  Results for a single camera and one clamp orientation are provided in Table~\ref{tab:synthvsreal}.}

\REV{The difference of success between the synthetic and the 'lab' dataset is around 20\% for the location estimation (prediction of $x$ and~$y$). Surprisingly, the network is better on the 'lab' and on the 'field' datasets for the orientation estimation (prediction of~$\theta$), compared to the synthetic dataset. We explain it by the variety of textures in the synthetic dataset acting as distractors whereas, in the 'lab' and 'field' datasets, the block edge are clearly visible and unambiguous.  Also, as all blocks have different sizes in the synthetic dataset, sometimes the short edge and the long edge of the block are of nearly similar length, leading to an orientation ambiguity, whereas the block dimensions in the real dataset are significantly different. As expected, the 'adv' dataset has poorer results.} 


\begin{table}
\caption{\REV{Success rate of fine pose estimation on the synthetic validation dataset and on the real evaluation datasets, for 1~camera and 1~clamp orientation.
}}
\centering
\label{tab:synthvsreal}       
\vspace{-2mm}
\begin{tabular}{l|c|c|c|c}
\noalign{\smallskip}\hline\noalign{\smallskip}
Dataset  & synthetic & 'lab' & 'field' & 'adv' \\
\noalign{\smallskip}\hline\noalign{\smallskip}
\% ($e_x$ $\leq$ 5\,mm)  &  \REV{$85.3$} & \OLD{$65.2$}$\REVV{62.8}$ & \OLD{$66.7$}$\REVV{61.1}$ & \OLD{$48.8$}$\REVV{47.5}$ \\
\% ($e_y$ $\leq$ 5\,mm)& \REV{$85.3$} &  \OLD{$63.8$}$\REVV{67.6}$ & \OLD{$63.6$}$\REVV{65.2}$ & \OLD{$47.8$}$\REVV{46.3}$ \\ 
\% ($e_\theta$ $\leq$ 2\textdegree)  & \REV{$68.4$} & \OLD{$83.7$}$\REVV{83.2}$ & \OLD{$77.7$}$\REVV{78.2}$ & \OLD{$63.3$}$\REVV{61.0}$\\
\noalign{\smallskip}\hline
\end{tabular}
\end{table}

\subsection{View aggregation}
\label{sec:resultsaggregation}

\begin{table}
\centering
\caption{Accuracy (mean and standard deviation of error) for fine estimation when aggregating different viewpoints (cameras 1, 2, 3) and different clamp orientations (P1 and P2) on the 'lab' dataset.}
\label{tab:multiview}       
\vspace{-2mm}
\begin{tabular}{l|c|c|c|c}
\multicolumn{5}{c}{$x$ error (mm)}   \\
\hline\noalign{\smallskip}
Setting & Camera 1 & Camera 2 & Camera 3 & All cameras   \\
\noalign{\smallskip}\hline\noalign{\smallskip}
Clamp P1 & \REV{$7.5 \pm 16.0$} & \REV{$8.2 \pm 16.6$} & \REV{$10.9 \pm 21.7$} & \REV{$3.4 \pm 7.6$} \\
Clamp P2 & \REV{$7.2 \pm 15.1$} & \REV{$6.7 \pm 11.0$} & \REV{$10.7 \pm 22.7$}  & \REV{$2.9 \pm 4.6$}\\
Both & \REV{$4.4 \pm 6.9$} & \REV{$4.1 \pm 6.2$} & \REV{$6.1 \pm 14.2$} & \REV{$\mathbf{2.3 \pm 1.8}$}\\
\noalign{\smallskip}\hline\noalign{\bigskip}

\multicolumn{5}{c}{$y$ error (mm)}   \\
\hline\noalign{\smallskip}
Setting  & Camera 1 & Camera 2 & Camera 3 & All cameras   \\
\noalign{\smallskip}\hline\noalign{\smallskip}
Clamp P1 & \REV{$6.7 \pm 14.6$}  & \REV{$7.1 \pm 15.6$} & \REV{$10.4 \pm 21.8$} & \REV{$3.0 \pm 5.4$} \\
Clamp P2 & \REV{$6.4 \pm 14.5$} & \REV{$5.5\pm 11.3$} &  \REV{$9.8 \pm 20.8$} & \REV{$2.8 \pm 4.3$} \\
Both & \REV{$4.4 \pm 7.7$} & \REV{$4.3 \pm 6.5$} &\REV{$6.5 \pm 15.0$} & \REV{$\mathbf{2.6 \pm 4.4}$} \\
\noalign{\smallskip}\hline\noalign{\bigskip}

\multicolumn{5}{c}{$\theta$ error (\textdegree)}   \\
\hline\noalign{\smallskip}
Setting  & Camera 1 & Camera 2 & Camera 3 & All cameras   \\
\noalign{\smallskip}\hline\noalign{\smallskip}
Clamp P1 & \REV{$1.4 \pm 3.7$} & \REV{$1.9 \pm 7.8$} & \REV{$2.2 \pm 8.0$}  & \REV{$0.8 \pm 0.6$} \\
Clamp P2 & \REV{$1.3 \pm 3.6$}& \REV{$1.4 \pm 3.9$} & \REV{$2.2 \pm 8.5$}  & \REV{$0.8 \pm 0.6$} \\
Both & \REV{$1.1 \pm 3.4$} & \REV{$1.1 \pm 4.4$}  & \REV{$1.3 \pm 5.6$} & \REV{$\mathbf{0.7 \pm 0.6}$} \\
\noalign{\smallskip}\hline
\end{tabular}
\end{table}

\REV{As indicated in Section~\ref{sec:classification}, the separate estimations of several cameras can be aggregated into a single global estimate.}
We evaluate in detail the interest of aggregating views from several cameras, also possibly considering together two different clamp orientations, at 90\textdegree, as shown on Figure~\ref{fig:gripor}.
We report\OLD{ed} our results for the fine estimation dataset in Table~\ref{tab:multiview}. This table allows two key observations.

First, by comparing the first three columns to the last one, one can see that unsurprisingly using several viewpoints improves performance. The boost in performance is actually quite significant.

Second, by comparing the first two lines of each table with the last one, one can notice that aggregating the predictions of two views from the same camera but with orthogonal clamp orientations boosts the results too. This is at first sight surprising as the two images are extremely similar, but it can be explained by the fact that the end part of the robot arm is almost symmetric and that estimating its orientation may be difficult from some orientations, as illustrated on Figure~\ref{fig:gripor}.

A little inconsistency may be observed in the estimation of~$y$ ($2.6 \pm 4.4$\,mm), which is not as accurate and robust as the estimation of~$x$ ($2.3 \pm 1.8$\,mm).  Although differences in general between $x$ and $y$ could be explained by the asymmetry of the robot base (see Figure~\ref{fig:teaser}) and the non-uniform range of positions of the cameras, located in the half space of positive $x$'s (see Figure~\ref{fig:acquisition}), the main reason here is actually that the network returns a totally wrong estimation (97\,mm error) for one of the 576 poses, whereas extreme mistakes are otherwise rare and never greater than 15\,mm.  This impacts $e_y$ for both the average (0.2\,mm) and standard deviation (more than 2\,mm).  Note that this single wrong estimation only affects the accuracy measures, not the success rates as it represents only 0.17\% of the poses.

Although not reported here, similar observations can be made about using different viewpoints for the coarse estimation, as well as when analyzing the percentage of images below a given\OLD{precision} accuracy instead of the average errors. These positive results validate our simple strategy to aggregate the predictions from each\OLD{view} camera and each clamp orientation. In the following, we only report aggregated results, unless otherwise mentioned.

\subsection{\OLD{Two}Three-step procedure and final grasping success rate}
\label{sec:resultthreesteps}

%

We presented a\OLD{two} three-step approach for estimating accurately the relative pose of a block with respect to the robot. We examine here whether it is realistic in terms of success rate  for the final grasping and if\OLD{both} all steps are really necessary.

\begin{table}
\centering
\caption{Success rate of coarse pose estimation on the real evaluation datasets, with 3 cameras and 1 clamp orientation.}
\label{tab:coarsesum}       
\vspace{-2mm}
\begin{tabular}{l|r|r|r}
\noalign{\smallskip}\hline\noalign{\smallskip}
Dataset  & \multicolumn{1}{c|}{'lab'} & \multicolumn{1}{c|}{'field'} &  \multicolumn{1}{c}{'adv'}  \\
\noalign{\smallskip}\hline\noalign{\smallskip}
\% ($e_x$ $\leq$ 60\,mm)  & \OLD{100.0} \REV{$99.8$} & \OLD{100.0} \REV{$100.0$} & \OLD{47.6} \REV{$84.7$}\\
\% ($e_y$ $\leq$ 60\,mm) & \OLD{100.0} \REV{$100.0$} & \OLD{100.0} \REV{$99.1$} & \OLD{53.4} \REV{$88.3$} \\
\% ($e_\theta$ $\leq$ 10\textdegree \OLD{degrees}) & \OLD{100.0} \REV{$99.8$} & \OLD{100.0} \REV{$99.1$} & \OLD{45.4} \REV{$88.7$} \\
\% ($e_x, e_y$ $\leq$ 60\,mm) & \REV{$99.8$} & \REV{$99.1$} & \REV{$77.5$} \\
\noalign{\smallskip}\hline
\end{tabular}
\vspace{2mm}
\end{table}

\REV{To check whether the approach makes sense, one can simply look at the success rate for the first two steps of the procedure. As can be seen from Table~\ref{tab:coarsesum}}, the first step, i.e., coarse pose estimation, has a success rate of 99.8\%. It is thus extremely reliable and almost always accurate enough to allow a subsequent fine pose estimation.
%
\REV{The second step of the procedure requires to locate the clamp in an image, to later crop a}\OLD{crop the input images to select the} region of interest for the fine localization. 
\REV{As the evaluation dataset does not include a ground truth of the exact position of the clamp in the images (see Section~\ref{sec:metrics}), we manually checked all the predictions of our trained network for localizing the center of the clamp and counted cases where the prediction was outside the bounding box of the clamp in the image.
In that respect, our second CNN correctly detects the clamp in 99.1\% of the pictures, which is also quite a high success rate. All in all, the success rate of the two steps prior to fine pose estimation is thus 98.9\%, which confirms that these preliminary steps do not significantly degrade the final estimation.}

Now to check whether the three-step procedure does improve accuracy over a single-step procedure, we\OLD{looked at} can compare the percentage of images with errors\OLD{bellow} below 5\,mm and 2\textdegree\ \OLD{for both setups}\REV{in two setups: one in which only a single broad view is considered, and one in which our three-step procedure performs a virtual close-up.} To obtain comparable results, and for this experiment only, both approaches are evaluated on the same input images, with the same resolution (1920\,$\times$\,1080 center-cropped to a square 1080\,$\times$\,1080) and with the same bin discretization (2\,mm, rather than 5\,mm used otherwise for coarse estimation).
Concretely, we test performance with the images where the clamp is positioned coarsely above the block, which is a necessary condition for the fine estimation step, but only a particular case of scene configuration for coarse estimation.  The results are reported in Table~\ref{tab:coarsevsfine}. It can be seen that the three-step procedure dramatically improves performance.

\begin{table}
  \caption{Success rate of pose estimation with a single-step procedure vs our three-step procedure on the 'lab' dataset.} 
\centering
\label{tab:coarsevsfine}       
\vspace{-2mm}
\begin{tabular}{l|c|c|c}
\noalign{\smallskip}\hline\noalign{\smallskip}
Procedure & Single-step & \multicolumn{2}{c@{~}}{Three-step} \\
\noalign{\smallskip}\hline\noalign{\smallskip}
         & 1 camera & 1 camera & 3 cameras \\      
Setup    & 1 clamp & 1 clamp & 2 clamp \\      
         &  position & position & positions \\      
\noalign{\smallskip}\hline\noalign{\smallskip}
\% ($e_x$ $\leq$ 5\,mm) & 20.5 & 61.1 & 90.8 \\
\% ($e_y$ $\leq$ 5\,mm) & 23.1 & 65.7 & 87.6 \\
\% ($e_\theta$ $\leq$ 2\textdegree) & 37.6 & 80.9 & 96.9 \\
\% ($e_x,e_y$ $\leq$ 5\,mm, $e_\theta$ $\leq$ 2\textdegree) & \hphantom{0}2.8 & 36.9 & 79.0 \\
\noalign{\smallskip}\hline
\end{tabular}
\end{table}

What makes the difference is that coarse pose estimation has to resize the image down to 256\,$\times$\,256 (a factor 4.2 on the length) to feed it into the network procedure, while fine pose estimation only resizes a 432\,$\times$\,432 crop centered on the clamp (a factor 1.7).  This resolution ratio of 2.5 between the two procedures naturally translates into a similar ratio for accuracy, and a considerable difference in the success rate for errors below 5\,mm.  
\OLD{this can be done easily using  LED system. It could also be done through 2D detection of the robot arm. To evaluate the sensitivity of our algorithm to this step of the procedure, we evaluated the performance depending on the crop size and report the results in Figure~\ref{fig:scale}. We can see that the performance decreases smoothly around an optimal crop scale, so that a 5\% error would barely affect accuracy, but a 10\% error would lead to a 20\% decrease in accuracy in the position estimate. This step is thus important, even if high precision is not critical for the application.}


\subsection{\REV{Comparison with other methods}}
\label{sec:baseline}

\REV{We found it difficult to compare our method to other approaches.  We tried various existing methods to detect cuboids, such as \citet{NIPS2012_4842}, but their performance on images of our evaluation dataset was too poor to be usable.  Likewise, usual corner detection was not reliable enough to locate blocks in the pictures.  Besides, block detection is only part of the problem as, besides camera intrinsic and extrinsic calibration, the robot also need to be registered in the camera frame for relative block positioning.}

\begin{figure}[t]
 \centering
 \begin{tabular}[width=\textwidth]{c c}
   \includegraphics[width=0.2\textwidth]{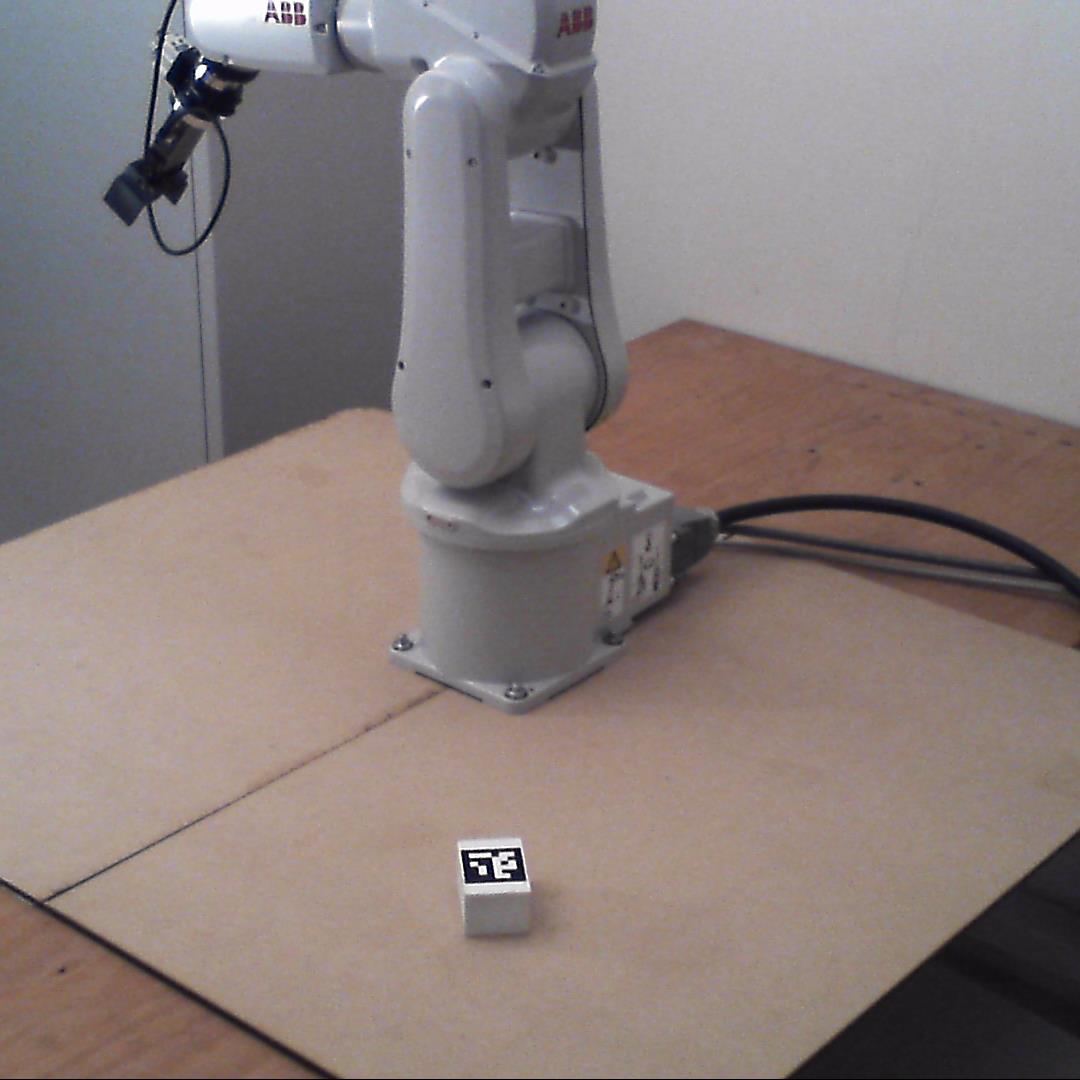} & \includegraphics[width=0.2\textwidth]{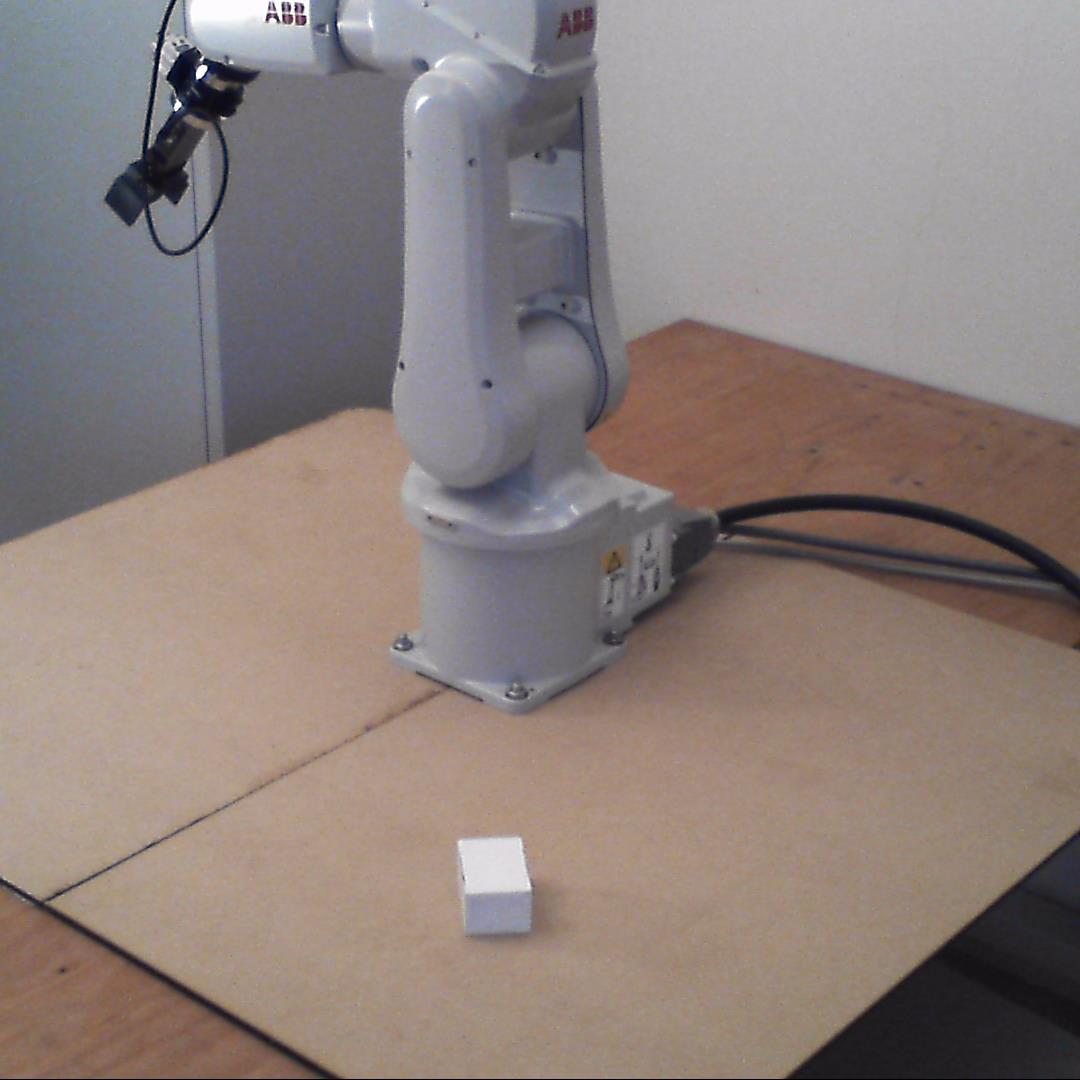} \\
   \small (a) &    \small (b)
 \end{tabular} \qquad
 \caption{Variant of the 'lab' dataset with identical poses but with a block having a fiducial marker on top of it}
 \label{fig:fiducial}
\end{figure}

\REV{To construct a simple baseline for comparison, we resorted to markers.  We built yet another dataset\OLD{, named 'mark',} which is identical to our 'lab' dataset except that a fiducial marker is added on the top of the block, as illustrated in Figure~\ref{fig:fiducial}.  On this dataset, we can detect the block position with the ArUco marker system \citep{GARRIDOJURADO20142280,GARRIDOJURADO2016481} as the 2D detection of the four corners of the marker performs well: the marker is correctly detected in 95.0\% of the pictures.}

\REV{As we use the same block, with a known height, for our entire dataset, and as the block is always laid flat on the table plane, we can use several known positions of the block, accurately placed on the table by the robot, to calibrate each fixed camera independently.  More precisely, we use these positions to determine the coefficients of the homography matrix relating the support plane to the image plane.  As the robot itself is used to place the block on the table at various locations, we also know the relative position of the robot with respect to the block.  After this calibration step, a 2D block detection in an image directly translates into a pose on the table plane, hence on a relative pose with respect to the robot base.
  Table~\ref{tab:baseline} compares our method to the results of pose estimation with the marker-based approach.  (The figures regarding our method differ from Table~\ref{tab:synthvsreal} because they do not only take into account fine pose estimation but the complete three-step procedure, with possible failures at coarse pose estimation and clamp detection.)
  As expected, the performance with markers is better than our method, with a single view.  However, our method reaches a comparable performance when aggregating the views of 3 cameras and 2 clamps orientations, without all the practical constraints of marker-based approaches.}

\begin{table}
\begin{REVregion}
\centering
\caption{\REV{Success rate of pose estimation with the marker-based method vs our three-step method on the 'lab' dataset.}} 
\label{tab:baseline}       
\vspace{-2mm}
\begin{tabular}{l|c|c|c}
\noalign{\smallskip}\hline\noalign{\smallskip}
\multicolumn{1}{l|}{Method}& \multicolumn{1}{c|}{Marker-based} & \multicolumn{2}{c}{Ours}  \\
\noalign{\smallskip}\hline\noalign{\smallskip}
 & 1 camera & 1 camera & 3 cameras \\
Setup  & 1 clamp & 1 clamp & 2 clamp \\
 & position & position & positions \\
\noalign{\smallskip}\hline\noalign{\smallskip}
\% ($e_x$ $\leq$ 5\,mm)  & $\REV{88.4}$ & $\REV{61.1}$ & $\REV{90.8}$ \\
\% ($e_y$ $\leq$ 5\,mm)  & $\REV{95.0}$ & $\REV{65.7}$ & $\REV{87.6}$ \\
\noalign{\smallskip}\hline
\end{tabular}
\end{REVregion}
\end{table}


\REV{This experiment emphasizes the fact that our approach does not solve a completely new task, or a task that could not be solved with existing tools. As the task is perfectly well defined, the performance of a "classical" approach mainly depends on the quality of camera calibration, and 2D/3D alignment algorithms for the robot and the block. Note however that camera calibration is time consuming, and we found it very difficult to find a robust and accurate 2D-3D detection and alignment algorithm. While using markers can make 2D-3D alignment easy, they also have strong practical constraints and could easily be stained, damaged or partially occluded in real-life scenarios, which would make them ineffective.
}


\subsection{Sensitivity to the environment}

The results we report\OLD{ed} in a laboratory environment show the potential of our method. \OLD{We wanted however}To go beyond these results and evaluate how robust our approach\OLD{was to} is, we consider more\OLD{complex} difficult settings, i.e., the more realistic 'field' conditions and the adverse 'adv' dataset. Results for the coarse pose estimation are reported in Table~\ref{tab:coarsesum} and, for fine pose estimation, in Tables \ref{tab:finesuccess} and~\ref{tab:fineaccur}.
As expected the results are better with the bare 'lab' environment. Interestingly, the results in more realistic 'field' environment are still satisfying, with\OLD{$100\%$} a success rate of \REV{$99.1\%$} for coarse estimation, and \REV{$70.2\%$}\OLD{the average error only slightly increasing around\OLD{4\,mm} \REV{2.6\,mm}} for fine estimation. However, in adverse conditions with pieces of papers that could easily be confused with the block (see Figure~\ref{fig:advdataset}), the performance drop is more dramatic, especially for coarse estimation where many distractors are visible, while the success rate of fine estimation \OLD{error}drops to \REV{$45.1\%$}. \OLD{increases to more than\OLD{7.5\,mm} \REV{3.8\,mm}.}\OLD{This result is not unexpected since the training images did not involve any distractor and a more adapted training dataset would be required to perform well under such conditions.}

\begin{table}
\centering
\caption{Success rate of fine pose estimation with 3 cameras and 2 clamp orientations on different (real) evaluation datasets.}
\label{tab:finesuccess}       
\vspace{-2mm}
\begin{tabular}{l|c|c|c}
\noalign{\smallskip}\hline\noalign{\smallskip}
Dataset   & 'lab' & 'field' &  'adv'  \\
\noalign{\smallskip}\hline\noalign{\smallskip}
\% ($e_x$ $\leq$ 5\,mm) & $\REV{91.8}$ & $\REV{89.5}$ & $\REV{73.4}$ \\
\% ($e_y$ $\leq$ 5\,mm) & $\REV{88.5}$ & $\REV{86.8}$ & $\REV{71.0}$ \\
\% ($e_\theta$ $\leq$ 2\textdegree) & $\REV{97.9}$ & $\REV{93.0}$ & $\REV{86.5}$ \\
\% ($e_x,e_y$ $\leq$ 5\,mm, $e_\theta$ $\leq$ 2\textdegree) &  $\REV{79.9}$ &  $\REV{70.2}$ &  $\REV{45.1}$ \\
\noalign{\smallskip}\hline
\end{tabular}
\end{table}

\begin{table}
\centering
\caption{Accuracy (mean and standard deviation of error) of fine pose estimation with 3 cameras and 2 clamp orientations on different (real) evaluation datasets.}
\label{tab:fineaccur}       
\vspace{-2mm}
\begin{tabular}{l|c|c|c}
\noalign{\smallskip}\hline\noalign{\smallskip}
Dataset   & 'lab' & 'field' &  'adv' \\
\noalign{\smallskip}\hline\noalign{\smallskip}
$e_x$ (mm)  & $\REV{2.3 \pm 1.8}$ & $\REV{2.3 \pm 1.8}$ & $\REV{4.0 \pm 4.4}$ \\
$e_y$ (mm) & $\REV{2.6 \pm 4.4}$ & $\REV{2.3 \pm 2.0}$& $\REV{4.0 \pm 5.2}$ \\
$e_\theta$ (\textdegree) & $\REV{0.7 \pm 0.6}$ & $\REV{0.9 \pm 0.7}$ & $\REV{1.1 \pm 0.9}$ \\
\noalign{\smallskip}\hline
\end{tabular}
\end{table}

\section{Conclusion}

We have introduced a new task of \emph{uncalibrated relative pose estimation} of an object with respect to a robot. The task can rely on a single view or on multiple views, possibly with different arm positions, and with a possible intermediate step for a more accurate pose estimation.  We have also constructed a rich dataset for the evaluation of this task.  Last, we have proposed a general method to perform this task, that provides a strong baseline. Indeed, our approach estimates the block pose with respect to the robot with an average location accuracy of \REVV{2.6\,mm} and an average orientation accuracy of \REVV{0.7\textdegree}, and these results in lab conditions degrade well in more realistic or adverse settings.  Given the small opening range of our test clamp, which requires a location error less than 5\,mm and an orientation error less than 2\textdegree, this translates into an overall success rate of 80\% for block grasping.  While these results are slightly lower than with methods relying on calibrated cameras and  robot, they show that relative pose estimation can be practically addressed in unfavorable settings where camera calibration is fragile or cannot be performed given the context.

One important strength of our method is that it can be learned with synthetic images only, without using a single real image for training, thus avoiding the expensive and time-consuming collection of training data.

Natural extensions of our approach include supporting different robots, different grasping tasks, and blocks of different shapes, at different heights on non-horizontal surfaces.

\bibliographystyle{spbasic}      
\bibliography{biblio}   

\begin{thebibliography}{44}
\providecommand{\natexlab}[1]{#1}
\providecommand{\url}[1]{{#1}}
\providecommand{\urlprefix}{URL }
\expandafter\ifx\csname urlstyle\endcsname\relax
  \providecommand{\doi}[1]{DOI~\discretionary{}{}{}#1}\else
  \providecommand{\doi}{DOI~\discretionary{}{}{}\begingroup
  \urlstyle{rm}\Url}\fi
\providecommand{\eprint}[2][]{\url{#2}}

\bibitem[{Aubry et~al(2014)Aubry, Maturana, Efros, Russell, and
  Sivic}]{aubry2014seeing}
Aubry M, Maturana D, Efros AA, Russell BC, Sivic J (2014) Seeing {3D} chairs:
  exemplar part-based {2D-3D} alignment using a large dataset of {CAD} models.
  In: Conference on Computer Vision and Pattern Recognition (CVPR), IEEE, pp
  3762--3769

\bibitem[{Chen et~al(2016)Chen, Wang, Li, Su, Wang, Tu, Lischinski, Cohen-Or,
  and Chen}]{chen2016synthesizing}
Chen W, Wang H, Li Y, Su H, Wang Z, Tu C, Lischinski D, Cohen-Or D, Chen B
  (2016) Synthesizing training images for boosting human {3D} pose estimation.
  In: 4th International Conference on 3D Vision (3DV), IEEE, pp 479--488

\bibitem[{Collet and Srinivasa(2010)}]{collet2010efficient}
Collet A, Srinivasa SS (2010) Efficient multi-view object recognition and full
  pose estimation. In: International Conference on Robotics and Automation
  (ICRA), IEEE, pp 2050--2055

\bibitem[{Collet et~al(2011)Collet, Martinez, and Srinivasa}]{collet2011moped}
Collet A, Martinez M, Srinivasa SS (2011) The {MOPED} framework: Object
  recognition and pose estimation for manipulation. The International Journal
  of Robotics Research (IJRR) 30(10):1284--1306

\bibitem[{Dalal and Triggs(2005)}]{dalal2005histograms}
Dalal N, Triggs B (2005) Histograms of oriented gradients for human detection.
  In: International Conference on Computer Vision and Pattern Recognition
  (CVPR), IEEE, vol~1, pp 886--893

\bibitem[{Dosovitskiy et~al(2015)Dosovitskiy, Fischer, Ilg, Hausser, Hazirbas,
  Golkov, van~der Smagt, Cremers, and Brox}]{dosovitskiy2015flownet}
Dosovitskiy A, Fischer P, Ilg E, Hausser P, Hazirbas C, Golkov V, van~der Smagt
  P, Cremers D, Brox T (2015) Flownet: Learning optical flow with convolutional
  networks. In: International Conference on Computer Vision (ICCV), IEEE, pp
  2758--2766

\bibitem[{Felzenszwalb et~al(2010)Felzenszwalb, Girshick, McAllester, and
  Ramanan}]{felzenszwalb2010object}
Felzenszwalb PF, Girshick RB, McAllester D, Ramanan D (2010) Object detection
  with discriminatively trained part-based models. IEEE Transactions on Pattern
  Analysis and Machine Intelligence (PAMI) 32(9):1627--1645

\bibitem[{Feng et~al(2014)Feng, Xiao, Willette, Mcgee, and
  Kamat}]{Feng2014TowardsAR}
Feng C, Xiao Y, Willette A, Mcgee W, Kamat VR (2014) Towards autonomous robotic
  in-situ assembly on unstructured construction sites using monocular vision.
  In: International Symposium on Automation and Robotics in Construction and
  Mining (ISARC)

\bibitem[{Fidler et~al(2012)Fidler, Dickinson, and Urtasun}]{fidler20123d}
Fidler S, Dickinson S, Urtasun R (2012) 3d object detection and viewpoint
  estimation with a deformable {3D} cuboid model. In: Advances in Neural
  Information Processing Systems (NIPS), pp 611--619

\bibitem[{Garrido-Jurado et~al(2014)Garrido-Jurado, Muñoz-Salinas,
  Madrid-Cuevas, and Marín-Jiménez}]{GARRIDOJURADO20142280}
Garrido-Jurado S, Muñoz-Salinas R, Madrid-Cuevas F, Marín-Jiménez M (2014)
  Automatic generation and detection of highly reliable fiducial markers under
  occlusion. Pattern Recognition 47(6):2280 -- 2292

\bibitem[{Garrido-Jurado et~al(2016)Garrido-Jurado, Muñoz-Salinas,
  Madrid-Cuevas, and Medina-Carnicer}]{GARRIDOJURADO2016481}
Garrido-Jurado S, Muñoz-Salinas R, Madrid-Cuevas F, Medina-Carnicer R (2016)
  Generation of fiducial marker dictionaries using mixed integer linear
  programming. Pattern Recognition 51:481 -- 491

\bibitem[{Girshick et~al(2014)Girshick, Donahue, Darrell, and
  Malik}]{girshick2014rich}
Girshick R, Donahue J, Darrell T, Malik J (2014) Rich feature hierarchies for
  accurate object detection and semantic segmentation. In: International
  Conference on Computer Vision and Pattern Recognition (CVPR), IEEE, pp
  580--587

\bibitem[{Glasner et~al(2011)Glasner, Galun, Alpert, Basri, and
  Shakhnarovich}]{glasner2011aware}
Glasner D, Galun M, Alpert S, Basri R, Shakhnarovich G (2011) Viewpoint-aware
  object detection and pose estimation. In: International Conference on
  Computer Vision (ICCV), IEEE, pp 1275--1282

\bibitem[{He et~al(2017)He, Gkioxari, Doll{\'a}r, and Girshick}]{he2017mask}
He K, Gkioxari G, Doll{\'a}r P, Girshick R (2017) Mask {R-CNN}. arXiv preprint
  arXiv:170306870

\bibitem[{Hejrati and Ramanan(2012)}]{hejrati2012analyzing}
Hejrati M, Ramanan D (2012) Analyzing {3D} objects in cluttered images. In:
  Advances in Neural Information Processing Systems (NIPS), pp 593--601

\bibitem[{Hoda{\v{n}} et~al(2016)Hoda{\v{n}}, Matas, and
  Obdr{\v{z}}{\'a}lek}]{hodavn2016evaluation}
Hoda{\v{n}} T, Matas J, Obdr{\v{z}}{\'a}lek {\v{S}} (2016) On evaluation of
  {6D} object pose estimation. In: European Conference on Computer Vision
  Workshops (ECCVw), Springer, pp 606--619

\bibitem[{Huttenlocher and Ullman(1990)}]{huttenlocher1990recognizing}
Huttenlocher DP, Ullman S (1990) Recognizing solid objects by alignment with an
  image. International Journal of Computer Vision (IJCV) 5(2):195--212

\bibitem[{LeCun et~al(1989)LeCun, Boser, Denker, Henderson, Howard, Hubbard,
  and Jackel}]{lecun1989backpropagation}
LeCun Y, Boser B, Denker JS, Henderson D, Howard RE, Hubbard W, Jackel LD
  (1989) Backpropagation applied to handwritten zip code recognition. Neural
  Computation 1(4):541--551

\bibitem[{Levine et~al(2016)Levine, Finn, Darrell, and Abbeel}]{levine2016end}
Levine S, Finn C, Darrell T, Abbeel P (2016) End-to-end training of deep
  visuomotor policies. Journal of Machine Learning Research (JMLR) 17(39):1--40

\bibitem[{Levine et~al(2018)Levine, Pastor, Krizhevsky, and
  Quillen}]{LevineISER2017}
Levine S, Pastor P, Krizhevsky A, Quillen D (2018) Learning hand-eye
  coordination for robotic grasping with deep learning and large-scale data
  collection. The International Journal of Robotics Research (ISER)
  37(4-5):421--436

\bibitem[{Lowe(1987)}]{lowe1987three}
Lowe DG (1987) Three-dimensional object recognition from single two-dimensional
  images. Artificial Intelligence 31(3):355--395

\bibitem[{Lowe(1999)}]{lowe1999object}
Lowe DG (1999) Object recognition from local scale-invariant features. In:
  Computer vision, 1999. The proceedings of the seventh IEEE international
  conference on, Ieee, vol~2, pp 1150--1157

\bibitem[{Massa et~al(2016{\natexlab{a}})Massa, Marlet, and
  Aubry}]{massa2016crafting}
Massa F, Marlet R, Aubry M (2016{\natexlab{a}}) Crafting a multi-task {CNN} for
  viewpoint estimation. In: 27th British Machine Vision Conference (BMVC)

\bibitem[{Massa et~al(2016{\natexlab{b}})Massa, Russell, and
  Aubry}]{massa2016deep}
Massa F, Russell BC, Aubry M (2016{\natexlab{b}}) Deep exemplar {2D-3D}
  detection by adapting from real to rendered views. In: International
  Conference on Computer Vision and Pattern Recognition (CVPR), IEEE, pp
  6024--6033

\bibitem[{Mundy(2006)}]{mundy2006object}
Mundy JL (2006) Object recognition in the geometric era: A retrospective. In:
  Toward category-level object recognition, Springer, pp 3--28

\bibitem[{Peng and Saenko(2017)}]{peng2017synthetic}
Peng X, Saenko K (2017) Synthetic to real adaptation with deep generative
  correlation alignment networks. arXiv preprint arXiv:170105524

\bibitem[{Peng et~al(2015)Peng, Sun, Ali, and Saenko}]{peng2015learning}
Peng X, Sun B, Ali K, Saenko K (2015) Learning deep object detectors from {3D}
  models. In: International Conference on Computer Vision (ICCV), IEEE, pp
  1278--1286

\bibitem[{Pepik et~al(2012)Pepik, Stark, Gehler, and
  Schiele}]{pepik2012teaching}
Pepik B, Stark M, Gehler P, Schiele B (2012) Teaching {3D} geometry to
  deformable part models. In: International Conference on Computer Vision and
  Pattern Recognition (CVPR), IEEE, pp 3362--3369

\bibitem[{Pepik et~al(2015)Pepik, Benenson, Ritschel, and
  Schiele}]{pepik2015holding}
Pepik B, Benenson R, Ritschel T, Schiele B (2015) What is holding back convnets
  for detection? In: 37th German Conference on Pattern Recognition (GCPR),
  Springer, no. 9358 in LNCS, pp 517--528

\bibitem[{Pinto and Gupta(2016)}]{PintoINCRA2016}
Pinto L, Gupta A (2016) Supersizing self-supervision: Learning to grasp from
  50k tries and 700 robot hours. In: International Conference on Robotics and
  Automation (ICRA), IEEE, Stockholm, Sweden, pp 3406--3413

\bibitem[{Richter et~al(2016)Richter, Vineet, Roth, and
  Koltun}]{richter2016playing}
Richter SR, Vineet V, Roth S, Koltun V (2016) Playing for data: Ground truth
  from computer games. In: European Conference on Computer Vision (ECCV),
  Springer, pp 102--118

\bibitem[{Roberts(1963)}]{roberts1963machine}
Roberts LG (1963) Machine perception of three-dimensional solids. PhD thesis,
  Massachusetts Institute of Technology (MIT)

\bibitem[{Ros et~al(2016)Ros, Sellart, Materzynska, Vazquez, and
  Lopez}]{ros2016synthia}
Ros G, Sellart L, Materzynska J, Vazquez D, Lopez AM (2016) The {SYNTHIA}
  dataset: A large collection of synthetic images for semantic segmentation of
  urban scenes. In: Internatgional Conference on Computer Vision and Pattern
  Recognition (CVPR), IEEE, pp 3234--3243

\bibitem[{Sadeghi and Levine(2018)}]{sadeghi2016cad}
Sadeghi F, Levine S (2018) {(CAD)2RL}: Real single-image flight without a
  single real image. In: Robotics: Science and Systems (RSS) Conference

\bibitem[{Schulman et~al(2015)Schulman, Levine, Abbeel, Jordan, and
  Moritz}]{schulman2015trust}
Schulman J, Levine S, Abbeel P, Jordan M, Moritz P (2015) Trust region policy
  optimization. In: 32nd International Conference on Machine Learning (ICML),
  pp 1889--1897

\bibitem[{Sermanet et~al(2014)Sermanet, Eigen, Zhang, Mathieu, Fergus, and
  LeCun}]{sermanet2013overfeat}
Sermanet P, Eigen D, Zhang X, Mathieu M, Fergus R, LeCun Y (2014) Overfeat:
  Integrated recognition, localization and detection using convolutional
  networks. In: International Conference on Learning Representations (ICLR)

\bibitem[{Shafaei et~al(2016)Shafaei, Little, and Schmidt}]{shafaei2016play}
Shafaei A, Little JJ, Schmidt M (2016) Play and learn: Using video games to
  train computer vision models. In: 27th British Machine Vision Conference
  (BMVC)

\bibitem[{Su et~al(2015)Su, Qi, Li, and Guibas}]{su2015render}
Su H, Qi CR, Li Y, Guibas LJ (2015) Render for {CNN}: Viewpoint estimation in
  images using {CNNs} trained with rendered {3D} model views. In: International
  Conference on Computer Vision (ICCV), IEEE, pp 2686--2694

\bibitem[{Sun and Saenko(2014)}]{sun2014virtual}
Sun B, Saenko K (2014) From virtual to reality: Fast adaptation of virtual
  object detectors to real domains. In: 25th British Machine Vision Conference
  (BMVC)

\bibitem[{Tobin et~al(2017)Tobin, Fong, Ray, Schneider, Zaremba, and
  Abbeel}]{tobin2017domain}
Tobin J, Fong R, Ray A, Schneider J, Zaremba W, Abbeel P (2017) Domain
  randomization for transferring deep neural networks from simulation to the
  real world. In: 30th International Conference on Intelligent RObots and
  Systems (IROS), IEEE/RSJ

\bibitem[{Tulsiani and Malik(2015)}]{tulsiani2015viewpoints}
Tulsiani S, Malik J (2015) Viewpoints and keypoints. In: International
  Conference on Computer Vision and Pattern Recognition (CVPR), IEEE, pp
  1510--1519

\bibitem[{Vazquez et~al(2014)Vazquez, Lopez, Marin, Ponsa, and
  Geronimo}]{vazquez2014virtual}
Vazquez D, Lopez AM, Marin J, Ponsa D, Geronimo D (2014) Virtual and real world
  adaptation for pedestrian detection. IEEE Transactions on Pattern Analysis
  and Machine Intelligence (PAMI) 36(4):797--809

\bibitem[{Wu et~al(2016)Wu, Xue, Lim, Tian, Tenenbaum, Torralba, and
  Freeman}]{wu2016single}
Wu J, Xue T, Lim JJ, Tian Y, Tenenbaum JB, Torralba A, Freeman WT (2016) Single
  image {3D} interpreter network. In: European Conference on Computer Vision
  (ECCV), Springer, pp 365--382

\bibitem[{Xiao et~al(2012)Xiao, Russell, and Torralba}]{NIPS2012_4842}
Xiao J, Russell B, Torralba A (2012) Localizing {3D} cuboids in single-view
  images. In: Advances in Neural Information Processing Systems (NIPS), Curran
  Associates, Inc., pp 746--754

\end{thebibliography}

\appendix

\section{Training Dataset Based on Synthetic Images}\label{sec:syntheticdataset}

\begin{figure}[t]
\centering
\includegraphics[width=0.75\columnwidth]{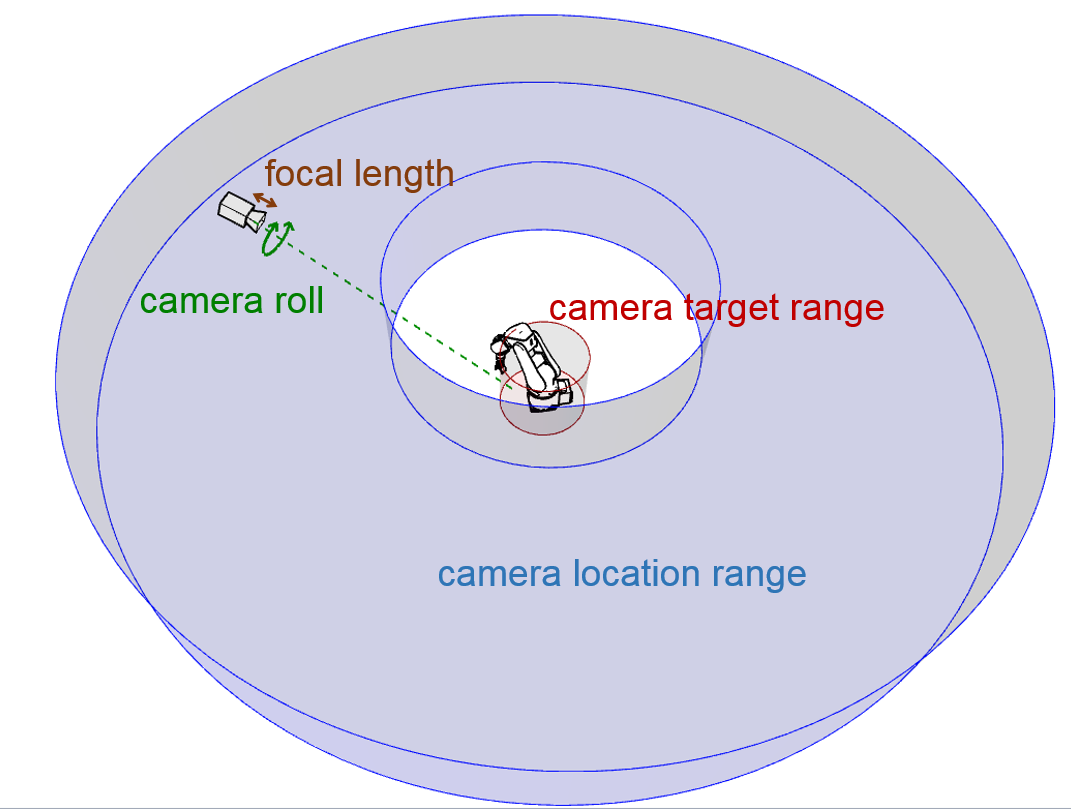}
\caption{Representation of some parameters defining a synthetic configuration of the relative pose between the robot and a camera}
\label{fig:camera}       
\end{figure}

\REV{As introduced in Section~\ref{learnfromsynth}, we generated three synthetic datasets for training a network for each subtask:}
\REV{
 \begin{enumerate}
  \item robot and block in random pose, for coarse estimation,
  \item robot with clamp in random vertical pose pointing downwards, for 2D tool detection,
  \item close-up of vertical clamp and random block nearby, for fine estimation.
 \end{enumerate}}
 \REV{We created a simple room model where a robot is placed on\OLD{a table} the floor and a cuboid block is laid nearby. The robot we experimented with is an IRB120 from ABB company, for which we have a 3D model. We also have a 3D model of the clamp.  However, we did not model the cables on the robot base nor on the clamp (compare, e.g., Figure~\ref{fig:zones} to Figure~\ref{fig:testimage}). We considered configurations that are similar to what can be found in the evaluation dataset, although with slightly greater variations for robustness. The actual randomization for the generation of images is as follows:}
\begin{itemize}[topsep=3pt]
\item \OLD{the table dimension, which varies randomly between 0.7 and 2\,m,} The size of the room is 20\,m\,$\times$\,20\,m so that the walls are visible on some images. \OLD{\TODO{Instead, we have to say a few things about the size of the room, as the walls are visible on some images.}}
\item The robot base (20\,cm\,$\times$\,30\,cm) orientation and position are sampled randomly.\OLD{\TODO{Does it make any sense to say the robot orientation and position is random when the camera position is random too?  It would only make sense with there was other objects, such as a wall, which is visible on a few pictures.}} (The robot height is around 70\,cm and the arm length around 50\,cm.)\OLD{\TODO{Introduce here, between parentheses, the basic dimensions of the robot, such as base sizes and arm length, otherwise the following numbers make little sense}}
\item The orientations (angles) of the robot joints are sampled randomly among all possible values, except for the clamp 2D detection task and the fine estimation task, where the arm extremity is placed vertically on top of the \OLD{table}floor.
\item Each dimension of the cuboid block\OLD{dimension, each dimension being} is sampled randomly between\OLD{1} 2.5 and 10\,cm.
\item The block is laid flat on the floor with an orientation and a position sampled randomly in the attainable robot positions for the coarse estimation task or in a 12\,cm square below the clamp for the fine estimation task.
\item All the textures,\OLD{which where} for the floor, robot and block are sampled among \OLD{59}69 widely different texture images\OLD{with a random scale factor}.
\item The camera center\OLD{positions, that are} is sampled randomly at a height between 70 and 130\,cm from the floor in a cylindrical sleeve of minimum radius 1\,m and maximum radius 2.8\,m, centered on the robot\OLD{regions we defined so that the robot is visible, between 1 and 4\,m from the camera}, as illustrated in Figure~\ref{fig:camera}.
\item For the coarse estimation setting (wide views), the camera target is sampled in a cylinder of 30\,cm radius and 50\,cm height around the robot base. 
\item For the fine estimation setting (close-ups), the target is the center of the clamp, with a small random shift. \OLD{\TODO{small shift? in m?}}
\item The camera is rotated around its main axis (line between camera center and camera target) with an angle sampled randomly between -8 and +8 degrees.
\item The camera focal length is randomly sampled between 45\,mm and 75\,mm for an equivalent sensor frame size of 24\,mm\,$\times$\,24\,mm.
\item Synthetic images are 256\,$\times$\,256 pixels. \OLD{The crop size for the fine estimation task is 40\% of the image width before scaling to 256\,$\times$\,256 pixels. \OLD{If the crop goes beyond the image border, it is recentered to fit in the image.}}
\end{itemize}

\OLD{We filter out of this dataset the synthetic images where more than 33\,\% the block is occluded.
The pictures were generated with the free-form modeler software Rhinoceros 3D and its Grasshopper editor.} 
\REV{ The pictures were generated with the Unreal Engine 4 game engine.}
The dataset we created for coarse estimation consists of approximately\OLD{360k} 420k images (examples can be seen on Figure~\ref{fig:synthcoarse}), the one for 2D clamp detection consists approximately  of 2800k images (examples can be seen on Figure~\ref{fig:synthclamp}) and the dataset for fine estimation consists approximately of\OLD{300k} 600k images (examples can be seen on Figure~\ref{fig:synthfine}). We used 90\% of the images for training and the remaining 10\% for validation. \OLD{In addition to the standard random crop data augmentation, we performed random 2D rotations up to 4 degrees.  This provides robustness to small actual camera rotations as well as to some perspective effects depending on focal length and distance to objects.}

\section{Evaluation Dataset Based on Real Images}\label{sec:evaluationdtataset}

\REV{%
As explained in Section~\ref{sec:realdata}, our evaluation dataset actually divides in three parts, corresponding to different settings, illustrated on Figure~\ref{fig:datasets}:
\begin{enumerate}[topsep=3pt,itemsep=3pt]
\item in the 'lab' dataset, the robot and the block are on a table with no particular distractor or texture,
\item in the 'field' dataset, the table is covered with dirt, sand and gravels, making the flat surface uneven,
\item in the 'adv' dataset, the table is covered with pieces of paper that can be confused with cuboid blocks.
\end{enumerate}}
We use\OLD{d} the robot itself to accurately move the block to random poses, which provides an reliable measure of its relative position and orientation, for each configuration.\OLD{A slight notch in the block ensures there is no drift as the robot repeatedly picks and moves the block.} \REV{In practice, the block can slowly drift from its expect position as the robot repeatedly picks it, moves it and puts it down.  To ensure there is no drift, the block position is checked and realigned every ten position.} Because of the limited stroke of the clamp, we considered only a single block sizes\OLD{, namely}: 5\,cm $\times$ 8\,cm $\times$ 4\,cm. 
Note however that our method does not\OLD{leverage} exploit this information\OLD{, and}; we believe a robust method should be able to process a wide range of block shapes. \REV{As we want to model situations where the robot can pick a block, we have to restrict the reach of the arm to a range for which the tool can be set vertically above the block, i.e., 0.505\,m.}


We collected images from 3 cameras at various viewpoints, looking at the scene slightly from above.  To make sure the block is visible on most pictures, we actually\OLD{sampled} considered successively different regions of the experiment table for sampling block poses\OLD{independently}, moving the cameras to ensure a good visibility for each region. The cameras are moved manually without any particular care. The distance between a camera and the block is typically between 1 and\OLD{4} 2.5\,m.  The maximum angle between the left-most and the right-most camera is typically on the order of 120 degrees.  This setting is illustrated on Figure~\ref{fig:acquisition}.

For each block position with respect to the robot base, we consider\OLD{d} two main articulations of the robot arm: a random arm configuration for the coarse location subtask, and an articulation where the clamp is vertical, pointing downward and positioned next to the block for the fine location subtask.  In the latter case,\OLD{more precisely,} we positioned the clamp\OLD{15\,cm} 150\,mm above the table surface, at a random horizontal position in a\OLD{6\,cm} 120\,mm square around the block.

We actually recorded two clamp orientations along the vertical axis: a first random orientation, and then a second\OLD{image} orientation where the clamp is rotated by 90 degrees (see Figure~\ref{fig:gripor}).  As the clamp orientation, with respect to which the fine estimation is defined, can be hard to estimate for some configurations, using two orientations allows more accurate predictions.

\OLD{For the fine estimation subtask, we also need to identify a relevant zone of the image to focus on, i.e., a 2D location.}
\OLD{and a size/scale.  Although we know the location of the block \OLD{w.r.t.} with respect to the robot, we cannot deduce the relevant zone in a image because the cameras are not calibrated.  We thus resort to a simple hardware device.  We placed a small LED at the end of the robot arm, close to the clamp, and we turn it on and off, as illustrated on Figure~ref{fig:led}, identifying its location as the luminous peak.  Such a system can easily be added on a robot arm, or replaced by any standard 2D localization method.  This defines a point in the images.  As for the region scale, we defined it manually for each quadrant around the robot and each camera position.  An accurate scale could be estimated with a slightly more complex LED system, or with a 2D localization method.}


\OLD{\REV{We do not have a ground truth of the exact position of the clamp in images for fine estimation of the relative block pose. (We would have had to run a full camera and robot calibration.)  Instead, for our experiments, we manually check predictions for localizing the center of the clamp and count cases where the prediction is outside the bounding box of the clamp in the image (see Section~\ref{sec:resultthreesteps}).}}

\OLD{For each of these variants,}In total, we considered approximately\OLD{800} 1300 poses (positions and orientations) of the robot and the block together.  
This lead to a total of approximately\OLD{29,000} 12,000 images, of size 1080\,$\times$\,1080 for wide views and 432\,$\times$\,432 for close-ups with corresponding ground-truth relative position and orientation\OLD{, that}. 
The cameras used are eLight full HD 1080p webcams from Trust. Camera intrinsics were not available nor estimated. Nevertheless, the focal length was roughly determined to be about 50\,mm and, in the synthetic images, the camera focal length was randomly set between 45\,mm and 75\,mm (see Appendix~\ref{sec:syntheticdataset}). 

Datasets and 3D models are available from \url{imagine.enpc.fr/~loingvi/unloc}.

\section{\REV{Network Architecture Details}}\label{sec:netdetails}

We define here what are the bins for the three different networks, addressing the three different subtasks.

\REV{The number of bins for the last layer of the coarse estimation network depends on the bin size and on the maximum range of the robotic arm with the tool maintained vertically, i.e., 0.505\,m (see Appendix~\ref{sec:evaluationdtataset}).} \REV{In practice, we defined bins of 5\,mm for the coarse estimation and 2\,mm for the fine estimation, both visualized by the fine red grid in Figure~\ref{fig:zones} and~\ref{fig:zoom}. These pictures also give a sense of how accurate the localization is compared to the sizes of the block and of the robot. For the angular estimation, we used bins of 5 and 2 degrees respectively. For the clamp detection network, we defined bins whose size is 2\% of the picture width.}
\REV{Since as stated above, we predict each dimension separately, this leads to 202 bins for $x$ and $y$ and 36 bins for $\theta$ for the coarse localization network, 60 bins for $x$ and $y$ and 90 bins for $\theta$ for the fine localization network and 50 bins for $x$ and $y$ for the clamp localization network.}

%
%

\end{document}